\documentclass[sigconf]{acmart}

\usepackage{xcolor,soul}
\usepackage{graphicx}

\usepackage{amssymb}%

\usepackage{pifont}%
\usepackage{hyperref}
\usepackage{enumitem}
\usepackage{subcaption}
\usepackage{booktabs}
\usepackage{multirow}

\newcommand{\tool}{\texttt{gaugeNN}}

\AtBeginDocument{%
  \providecommand\BibTeX{{%
    \normalfont B\kern-0.5em{\scshape i\kern-0.25em b}\kern-0.8em\TeX}}}

\newif\ifacm
\acmfalse

\ifacm
\newcommand{\cready}[1]{#1}

\copyrightyear{2021}
\acmYear{2021}
\setcopyright{acmcopyright}
\acmConference[IMC '21]{ACM Internet Measurement Conference}{November 2--4, 2021}{Virtual Event, USA}
\acmBooktitle{ACM Internet Measurement Conference (IMC '21), November 2--4, 2021, Virtual Event, USA}
\acmPrice{15.00}
\acmDOI{10.1145/3487552.3487863}
\acmISBN{978-1-4503-9129-0/21/11}

\else
\newcommand{\cready}[1]{{#1}}
\usepackage{background}
\backgroundsetup{angle=0,
    scale=1,
    color=black,
    firstpage=true,
    position=current page.north,
    hshift=0pt,
    vshift=-20pt,
    contents={\ifnum\value{page}=1 PREPRINT: Accepted at the ACM Internet Measurement Conference (IMC), 2021 \else \fi}
}

\setcopyright{none}
\renewcommand\footnotetextcopyrightpermission[1]{} %

\fi

\begin{document}

\title[Smart at what cost? Characterising Mobile DNNs in the wild]{Smart at what cost? Characterising Mobile \\Deep Neural Networks in the wild}

\author{
{Mario Almeida*$^\dagger$, Stefanos Laskaridis*$^\dagger$, \\Abhinav Mehrotra$\dagger$, Lukasz Dudziak$\dagger$, Ilias Leontiadis$^\dagger$, Nicholas D. Lane$^{\dagger,\ddagger}$}}

\affiliation{\institution{$^\dagger$Samsung AI Center, Cambridge\hspace{+0.75cm}$^\ddagger$University of Cambridge}{\Small\textit{{* Indicates equal contribution.}}}}

\email{{mario.a, stefanos.l, a.mehrotra, l.dudziak, i.leontiadis, nic.lane}@samsung.com}

\authorwithoutinstuition{{Mario Almeida, Stefanos Laskaridis, Abhinav Mehrotra, Lukasz Dudziak, Ilias Leontiadis, Nicholas D. Lane}}

\renewcommand{\shortauthors}{Almeida, Laskaridis et al.}

\newif\ifcomment

 \commentfalse

\ifcomment
\newcommand{\stelios}[1]{\sethlcolor{lime}\hl{[\textbf{Stelios:} #1]}}
\newcommand{\steve}[1]{\sethlcolor{cyan}\hl{[\textbf{Steve:} #1]}}
\newcommand{\manote}[1]{\sethlcolor{pink}\hl{[\textbf{Mario:} #1]}}
\newcommand{\il}[1]{\sethlcolor{yellow}\hl{[\textbf{Ilias:} #1]}}
\newcommand{\nic}[1]{\sethlcolor{green}\hl{[\textbf{Nic:} #1]}}
\newcommand{\cut}[1]{\sethlcolor{light_red}\hl{[#1]}}
\newcommand{\abh}[1]{\sethlcolor{yellow}\hl{[\textbf{Abhi:} #1]}}
\newcommand{\luk}[1]{\sethlcolor{magenta}\hl{[\textbf{Lukasz:} #1]}}
\else
\newcommand{\stelios}[1]{}
\newcommand{\steve}[1]{}
\newcommand{\il}[1]{}
\newcommand{\nic}[1]{}
\newcommand{\manote}[1]{}
\newcommand{\cut}[1]{}
\newcommand{\abh}[1]{}
\fi

\begin{abstract}
With smartphones' omnipresence in people's pockets, \cready{Machine Learning (ML)} on mobile is gaining traction as devices become more powerful. 
With applications ranging from visual filters to voice assistants, intelligence on mobile comes in many forms and facets. 
However, \cready{Deep Neural Network (DNN)} inference remains a compute intensive workload, with devices struggling to support intelligence at the cost of responsiveness.
On the one hand, there is significant research on reducing model runtime requirements and supporting deployment on embedded devices. On the other hand, the strive to maximise the accuracy of a task is supported by deeper and wider neural networks, making mobile deployment of state-of-the-art DNNs a \mbox{moving target.}

In this paper, we perform the first holistic study of DNN usage in the wild in an attempt to track deployed models and match how these run on widely deployed devices. To this end, we analyse over 16k of the most popular apps in the Google Play Store to characterise their DNN usage and performance across devices of different capabilities, both across tiers and generations.
Simultaneously, we measure the models' energy footprint, as a core cost dimension of any mobile deployment. 
To streamline the process, we have developed \tool{}, a tool that automates the deployment, measurement and analysis of DNNs on devices, with support for different frameworks and platforms.
Results from our experience study paint the landscape of deep learning deployments on smartphones and indicate their popularity across app developers. Furthermore, our study shows the gap between bespoke techniques and real-world deployments and the need for optimised deployment of deep learning models in a highly dynamic and heterogeneous ecosystem.

\end{abstract}
\ifacm
\vspace{-0.8cm}
\fi

\maketitle

\vspace{-0.2cm}
\section{Introduction}

The recent popularity of Deep Neural Networks (DNNs) has seen them being applied to myriads of areas, from computer vision \cite{He_2016} to speech recognition \cite{chan2015listen} and machine translation \cite{sutskever2014sequence}.
DNNs are no longer only being deployed in datacenters \cite{facebook_datacenter_2018},
as they have found their way into mobile devices, ranging from IoT devices to flagship smartphones and self-driving cars. 
In fact, large part of what makes smartphones smart, can be attributed to the ever-increasing support for machine learning, be it in the form of camera optimisations, intelligent assistants or text predictions.

While DNNs have become more and more accurate, this was frequently at the expense of an increased number of parameters\cready{, energy consumption} and computational load \cite{almeida2019embench,Simonyan14c,Huang2017a, He_2016}, often resulting in poor performance 
on resource-restricted mobile and embedded devices \cite{10.1109/MICRO.2018.00022,48305,almeida2019embench}. %

To address these challenges, there has been significant research towards mobile-specific DNN optimisations. Firstly, researchers have designed various mobile-specific architectures either manually~\cite{howard2017mobilenets,hapi2020iccad} or automatically, through Network Architecture Search  (NAS)~\cite{Tan_2019_CVPR}. Secondly, numerous works have looked into reducing computation through weight sparsification and pruning~\cite{lee2019snip} and quantisation~\cite{han2016deep}.
Thirdly, kernel optimisations have been proposed for mobile SoCs \cite{marques_mlsys2020}. Last but not least, inference offloading is an alternative approach where computation is partly or wholly outsourced to a remote endpoint for faster results \cite{Kang2017, spinn2020mobicom}.

At the same time, recent developments on mobile SoCs enable smartphones to support higher DNN computational throughput at a lower energy budgets \cite{wang_2020,ai_benchmark_2019}, either through heterogeneous multi-core processors (e.g. ARM big.LITTLE \cready{and} DynamIQ) or through specialised hardware (e.g. DSPs \cready{and} NPUs). However, the device ecosystem remains very heterogeneous, ranging from cheaper devices with older processors to flagship devices with dedicated processing units.  As a result, it is extremely hard for developers to assess the performance and optimise their DNN \cready{models} for each possible device tier \cite{facebook2019}. 

In this work, we attempt to measure what the actual mobile ML landscape looks like in the wild by studying real-world DNNs, as deployed with the most popular applications of the Google Play Store. 
Our goal is to examine whether real-life deployments follow the state-of-the-art of ML research and identify performance bottlenecks over devices of different tiers and generations.
The gained experience will provide insights on the system and model-level optimisations required to push the current frontier of mobile intelligence. 
In particular, we make the following contributions:
\begin{itemize}[leftmargin=*,topsep=0pt]
\item We design a system, named \tool{}, that automates the extraction, analysis and benchmarking of DNN models found in the most popular apps in the wild.
\item Using \tool{} we analyse over 16k (33k across two snapshots) Google Play Store apps with respect to their DNN models. We characterise these models in terms of their usage, architecture, layer operations and optimisations as well as external \cready{cloud-based} DNN API calls. 
\item We compare our latest snapshot with a previous version of \cready{the Google Play most popular apps} 12 months ago and comment on the trajectory of DNN mobile penetration in the past year.
\item We perform a runtime measurement of hundreds of these DNN \cready{models} across heterogeneous devices of different capabilities to further characterise these models in terms of their achieved latency and energy consumption.
\item We analyse model and system-level optimisations supported by publicly available toolsets and provide an overview of the current DNN optimisation landscape available to developers and practical guidelines for improving the development and deployment of future DNNs.
\end{itemize}

\section{Research Questions \& Results}

With our study, we aim to answer the following Research Questions (RQ) that arise:
\begin{itemize}[leftmargin=*,label={}]
    \item \textbf{RQ\#1}: \textit{Given the forefront of ML research and the multitude of tools and devices in the wild,  what kind of models are being deployed in mobile apps and utilised by developers and for which tasks?}
    \item \textbf{RQ\#2}: \textit{In a highly heterogeneous ecosystem of smartphones, how are these models deployed and are they able to perform efficiently across different targets and tasks?}
    \item \textbf{RQ\#3}: \textit{What are common model and system-level optimisations being used to make inference in the wild faster on smartphones? Can they be improved?}
\end{itemize}

\noindent
\textbf{Results:}
Our results indicate that mobile developers choose to deploy simple off-the-shelf models on-device, potentially pretrained or fine-tuned for targeting different tasks, and often rely on cloud offloading to support larger tasks. This minimises the burden to the app developer and cashes upon existing models widely available.
Furthermore, we witness that devices of different tiers and generations have widely varying performance over the benchmarked models, with the low-tier devices being significantly slower in DNN-based tasks. 
When it comes to performance per watt, we notice a general trajectory of devices getting incrementally more efficient from generation to generation, with SoCs integrating more and more specialised hardware in the die. However, the same trajectory cannot be traced on battery technology, which remains largely the same and mainly varies depending on the device's form factor.
Last, we have observed that off-the-shelf model-level optimisations deployed with major frameworks more often than not do not result to latency or memory benefits during inference, but are focused on compressibility of the model. Simultaneously, SoC vendor-specific tools offer a significant benefit in runtime, at the expense of generality of the deployed models.
Still, we found no significant evidence of target-specific model deployment in the wild.

\section{Methodology}
\begin{figure}[t]
\centering
\includegraphics[width=.85\columnwidth]{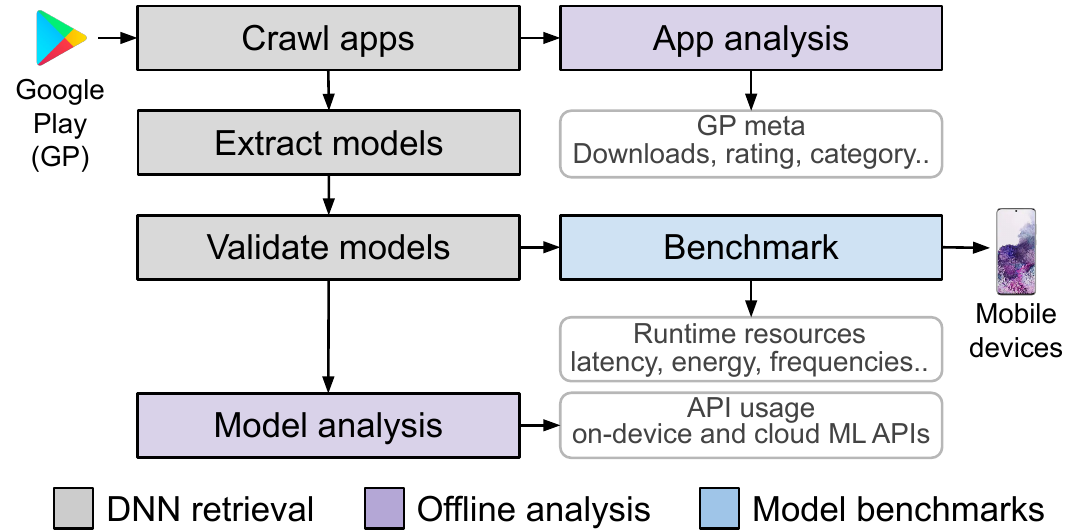}
\vspace{-0.4cm}
\caption{Workflow of \tool{}.}
\vspace{-0.4cm}
\label{fig:workflow}
\end{figure}

To fulfil these diverse characterisation goals, we employ the three step methodology depicted in Fig.~\ref{fig:workflow}.
First, we crawl the Google Play Store to find the DNN models from within the most popular apps among mobile users and extract their associated ML models, validating them against certain rules (\textit{grey boxes}).
Second, we perform a device-agnostic app and model analysis (\textit{purple boxes}).
Specifically, we look at the app's store metadata, where the DNN is used, as well as the model's layers and operations.
Finally, we benchmark the models on different devices to analyse their performance upon deployment (\textit{blue box}).
To automate this process and analyse ML models at scale we designed \tool{}.
We describe below each component in greater detail.

\subsection{DNNs retrieval}
\label{sec:crawling}

The first step in our methodology is to find, extract and validate the DNNs from Google Play Store most \mbox{popular apps.}

\noindent
\textbf{App crawling.} First, \tool{} mimics the web API calls made from the Google Play store of a typical mobile device to crawl the Google Play Store. In these requests, both the user-agent and locale headers are defined, which determine the variant of the store and apps retrieved.
To perform the crawling, we fetch the list of the top free apps per category which returns a maximum of 500 apps. 
Additionally, \tool{} stores the store metadata for each app, including popularity, category, reviews, etc. in an ElasticSearch instance 
for quick ETL\footnote{\cready{Evaluate Transform Loop}} analytics and cross-snapshot investigations
(Sec.~\ref{sec:offline_eval}).

\noindent
\textbf{Model extraction.} Given the downloaded apps, \tool{} proceeds to extract the DNN models from each application's package.
Traditionally, Android applications are packaged in a zip file, i.e.~\texttt{apk}, which comes with the the Java/Kotlin ``bytecode'' along with resources used by the app (e.g. textures, images, fonts).
\texttt{Apk}s have a size limit of 100MB and files -- such as DNN weights -- can have a larger storage footprint. As a result,  Google Play allows additional content to be shared either with expansion files \cite{apk_expansion_files}
(\texttt{OBB}s) or through Android App Bundles through Play Asset Delivery \cite{android_app_bundles}
The former supplement the main \texttt{apk} file and are hosted and served by Google Play, whereas the latter offers the possibility of downloading assets on demand,  as needed for a given device.
\tool{} supports file extraction from i)~the base \texttt{apk}, ii)~expansion files (\texttt{OBB}s) and iii)~Android App Bundles, but does not track asset delivery outside of Google Play.
Extracted files are matched against a compiled list of 69 known DNN framework formats (listed in the Appendix) to identify potential DNN models.

\begin{figure}[t]
\centering
\includegraphics[width=0.42\textwidth,clip={0 0 0 10mm},clip]{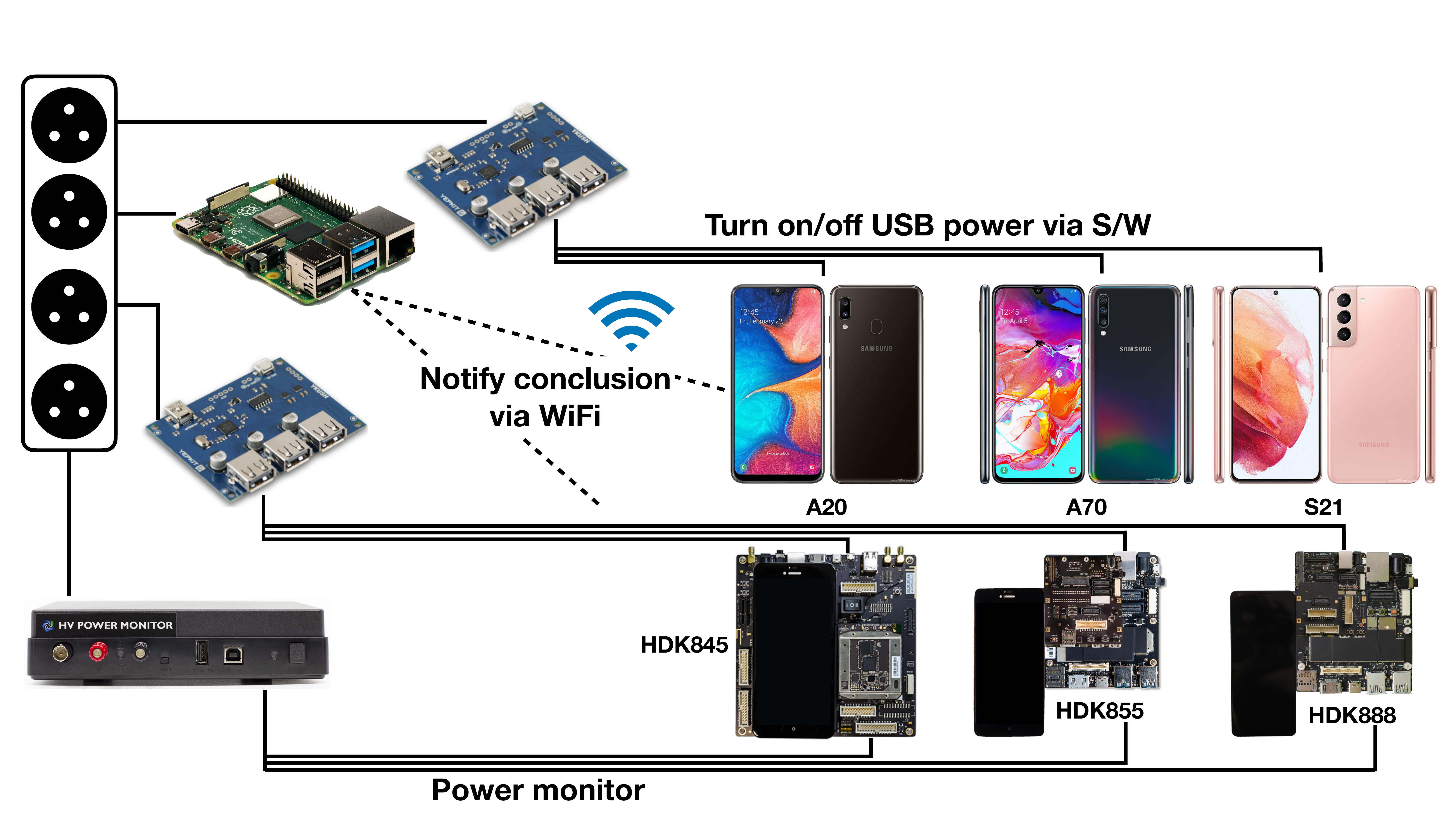}
\vspace{-0.4cm}
\caption{\tool{} benchmark platform.}
\vspace{-0.4cm}
\label{fig:energy}
\end{figure}

\noindent
\textbf{Model validation.}
Many models use generic file formats (e.g., protobuffer). Therefore, the number of candidate model files and extensions is quite large and benchmarking all prospective ones quickly becomes computationally prohibitive at scale. 
Therefore, inspired by \cready{the open-source} Netron \cite{netron_github} tool
, \tool{} employs a lightweight -- framework and format specific --validation process to remove files that are not DNN models.
This validation consists of checking the binary signature of the file for the presence of specific identifiers that a \mbox{framework uses.}
For example, for \texttt{TFLite}, we know that the FlatBuffer files representing models include specific headers at certain positions of the binary file, thus we check for the existence of e.g.~ the string ``\texttt{TFL3}'' there.

On the downside, encrypted and obfuscated models do not match such validation rules and are not extracted in our analysis. Moreover, models downloaded on demand by the application outside of the official Google Play distribution mechanisms are omitted from our benchmarks. However, we do track applications using such models indirectly by means of library inclusion in the application code and native libraries, even without explicitly analysing the models. \cready{The native code detection follows the methodology of Xu et al.}~\cite{xu2019first}.

\subsection{Offline DNN analysis}
\label{sec:methodology_offline}

After collecting the top apps from each category, we  analyse the usage of Deep Neural Networks in the wild.
Apps can use DNN models in different ways; i) they can execute the  models on-device or ii) offload the computation to external resources (e.g.~cloud providers). 

\noindent
\textbf{In-app DNN models.} 
After identifying the model files within an application, \tool{} extracts their DNN architecture
either by parsing directly the file, or by using the associated framework's interpreter. A DNN model is typically represented as a DAG\footnote{\cready{Directed Acyclic Graph}}, where layers are represented by vertices and data flows by edges.
By going through each model's graph, \tool{} registers the type of layer, its parameters (weights) and operations in a trace-based manner and uses this information to estimate the total operations\footnote{\cready{Model FLOPs are estimated as a function of the cumulative Multiply-Accumulate (MAC) operations performed by each of the model's layers.}} (\#FLOPs) and model size (\#parameters). Furthermore, we can later individually run these models and measure their inference latency, energy and memory footprint. 

\noindent
\textbf{DNN Cloud APIs.} Alternatively, applications might integrate ML functionality through cloud-backed APIs, by means of offloading inference to a remote endpoint.
To detect the usage of cloud-based DNN models, \tool{} inspects the app code to search for common DNN framework API calls.
Android apps are typically developed in Kotlin or Java and then compiled into \texttt{dex} format\cite{dalvik} and packaged within the app binary.
It is possible to extract this \texttt{dex} binary from the app package and decompile it into a human-readable (\texttt{smali}~\cite{smali}) format using the \texttt{apktool}~\cite{apktool}
to inspect the original code API calls. 
\tool{} automates the process of decompiling these binaries and performs string matching on the smali files to detect known cloud DNN framework calls.
In particular, \tool{} recognises calls to libraries belonging to Google FireBase~\cite{googlefirebase}, Google Cloud~\cite{googlecloud} and Amazon AWS ML services~\cite{awssdk}.

\begin{figure}[t]
\centering
\includegraphics[width=0.68\columnwidth]{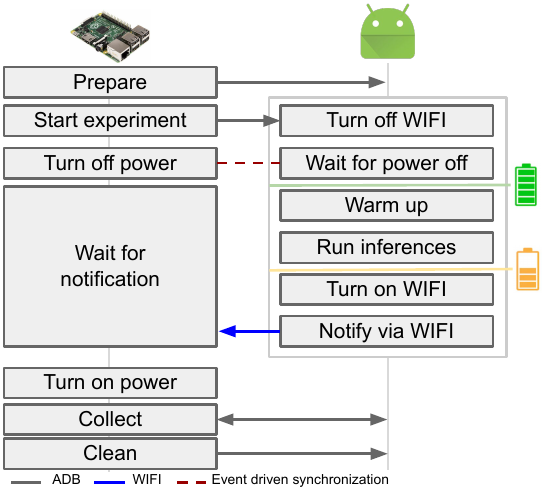}
\vspace{-0.4cm}
\caption{\tool{} benchmark workflow.}
\vspace{-0.45cm}
\label{fig:benchwork}
\end{figure}

\subsection{Model benchmarking}
\label{sec:benchmarking}

Next, we describe how \tool{} assesses the on-device run time and energy consumption of DNNs.

\noindent
\textbf{Devices.} To assess the performance of the deployed DNN models at runtime -- i.e. latency, energy, memory and CPU utilisation -- we deploy these models on the devices of Table~\ref{tab:specs}.
The devices of the first group represent three distinct tiers of smartphones (\textit{low} to \textit{high-end}) and showcase the performance across heterogeneous clients, while the development boards of the second group represent high-tier SoCs from different generations, whose open design allows us to measure energy consumption through cable probes connected to a Monsoon power monitor (Fig.~\ref{fig:energy}).

\noindent
\textbf{Benchmark workflow.} All benchmarks are written in native code and compiled for \texttt{aarch64} with Android NDK.
\tool{} adopts a master-slave architecture depicted in Fig.~\ref{fig:energy}. The server, where the models initially reside, is responsible for orchestrating the deployment and benchmarking of the models across client devices (phones), connected over USB. To control the power passthrough of mobile devices, we use a USB controller board \cite{ykush} that can programmatically disable data and power channels during measurements. This component was necessary, as connecting the device over USB charges it, interfering with the \mbox{energy measurements}.

The benchmarking workflow is depicted in Fig.~\ref{fig:benchwork}. Initially, the master (left side) pushes all the necessary dependencies to the device (right side) through \texttt{adb} and asserts the initial device state (WiFi and sensors are off, maximum screen timeout, etc).
The benchmark consists of an unattended, headless script that runs on the device upon disconnection of the USB power, controlled through the USB board.
This script is launched as a daemon process and performs the following tasks: 1)~It waits until the USB power is off; 2)~it runs a configurable amount of warmup inferences to remove cold cache outliers; 3)~it runs the actual benchmark inferences with a configurable inter-experiment sleep period; 4)~it turns on WiFi upon completion and communicates a TCP message through \texttt{netcat} to the server that the experiment is over.
Subsequently, the server re-enables the USB power, connects over \texttt{adb} and gathers the job results before cleaning up and launching the next job.

\begin{table}[t]
\small
\begin{tabular}{llll}
\hline
Model & SoC & RAM & Battery capacity \\ \hline
\multicolumn{4}{l}{\footnotesize{\textbf{Samsung devices}}} \\
A20 & Exynos 7884 & 4GB & 4000mAh  \\ 
A70 & Snapdragon 675 & 6GB & 4500mAh \\ 
S20 & Snapdragon 888 & 8GB & 4000mAh \\ 
\hline
\multicolumn{4}{l}{\footnotesize{\textbf{Qualcomm development boards}}} \\ 
Q845 HDK & Snapdragon 845 & 8GB & 2850mAh\\
Q855 HDK & Snapdragon 855 & 8GB & N/A \\
Q888 HDK & Snapdragon 888 & 8GB & N/A \\ \hline
\end{tabular}
\vspace{0.1cm}
\caption{Device specifications.}
\vspace{-.95cm}
\label{tab:specs}
\end{table}

\noindent
\textbf{Energy measurements.} 
Energy on open deck devices is measured via a Monsoon power monitor (\texttt{AAA10F}).
To prevent Android's battery saving mechanisms (e.g., Doze~\cite{doze}) killing background jobs when the screen goes off or scaling down the CPU frequency, we keep the phone screen on during the benchmark, by interfacing with the Android's Power Manager service. We also ensure that the screen is always in a similar state across devices, by developing an app that shows a black background. While the screen does incur extra energy consumption, this is measured and accounted for.

In the following sections, we present the findings of our experiments run with \tool{}. First, we present an offline analysis of the apps and models found from crawling the Google Play Store (Sec.~\ref{sec:offline_eval}) and then we move to runtime analysis of these models on devices (Sec.~\ref{sec:benchmarking}) and specific optimisations (Sec.~\ref{sec:optimisations}).

\label{sec:methodology}

\section{Dataset Collection \& Analysis}
\label{sec:offline_eval}
In this section, we attempt to find an answer to \textbf{RQ\#1} \cready{with regards to} DNN deployment in the wild. To this direction we first analyse %
our collected data with respect to 
the existence of DNN models in the top Google Play Store apps and their distribution to user devices.
Then we move to more specific model and app \textit{categorisation} and \textit{characterisation} and finally draw conclusions about the trajectory of ML mobile deployment from our temporal analysis results.

\vspace{-0.2cm}
\subsection{Datasets}

As shown in Table~\ref{tab:dataset} we collected two snapshots of the top free Google Play apps, on the $14^{th}$ of February 2020 and on the $4^{rd}$ of April 2021.
At these points in time, the Android devices represented $73.3\%$ and $72.19\%$ of the mobile OS market share \cite{statista_market_os, gs_market_os} respectively.
Data was collected from \cready{an UK-based account associated to a Samsung S10 (SM-G977B)}, downloading the most popular apps across all categories of the Google Play Store (up to 500 apps per category). This accounts for the top 0.6\% of total applications available in the store\footnote{Google Play Store is estimated to have 2.9M apps at the time of the \mbox{latest snapshot \cite{appbrain_stats}}}. 
In general, apps downloads tend to follow a power law distribution \cite{viennot2014measurement}. Therefore, the most popular apps are installed on most users' phones while the rest follow a long tail. While we could not scale a study of paid apps for monetary reasons, these account for a very small percentage of downloaded apps~\cite{viennot2014measurement}.
For the rest of the paper, we report on the latest Play Store snapshot, unless explicitly stated otherwise.

\vspace{-0.2cm}
\subsection{Model distribution to devices}
As described in Sec.~\ref{sec:crawling}, models in Android applications can be distributed post-installation (e.g. through \texttt{OBB}s or Asset Delivery). This allows developers to bypass the 100MB \texttt{apk} limit and to provide customised models for devices with different capabilities (e.g. devices with specified NPU).
To identify any models that are distributed post-installation, we downloaded all companion files and Google Play assets. \cready{We found} no models being distributed outside of the main \texttt{apk}.
\cready{Furthermore, we downloaded an extra snapshot with a three Android generations older device profile}\footnote{\cready{Samsung S7 edge -- SM-G935F, released in February'16, three years before the S10 5G.}}, and found no evidence of device-specific model customisation.

\noindent
\textbf{Observations:} \textit{Our results indicate that the functionality offered by Play Services to download device-specific models may be underutilised in the realm of mobile ML or that developers choose not to specialise their models per device SoC or model.
\cready{While specialising the model distribution per device target can be beneficial for performance and energy, it requires offline vendor-specific customisation of the model. Evidently, app developers seem to prefer generality of their deployment solutions, in line with} \cite{facebook2019}, \cready{and defer optimisation to middleware in the stack, such as NNAPI drivers or specific hardware delegates} \cite{ai_benchmark_2019}.}

\begin{table}[]
\small
\begin{tabular}{lrr}
\hline
                            & {Snapshot '20}                                 & Snapshot '21 \\ \hline
\textbf{Date}               & $14^{th}$ Feb. 2020                            & $4^{rd}$ Apr. 2021                             \\
\textbf{Total Apps}         & $16,964$                                       & $16,653$                                       \\
\textbf{Apps w/ frameworks} & $236(1.4\%)$                                   & $377 (2.3\%)$                                  \\
\textbf{Apps w/ models}     & $165 (1.0\%)$                                  & $342 (2.1\%)$                                  \\
\textbf{Total models}       & $821$                                          & $1,666$                                        \\
\textbf{Unique models}      & $129 (15.7\%)$                                 & $318 (19.1\%)$                                 \\ \hline
\end{tabular}
\vspace{0.1cm}
\caption{Dataset snapshots details.}
\vspace{-0.75cm}
\label{tab:dataset}
\end{table}

\vspace{-0.2cm}
\subsection{ML frameworks}

Next, we look into the models found per ML framework. 
Specifically, Fig.~\ref{fig:categories} depicts the number of models successfully extracted, validated and benchmarked, per category and ML framework.
These models represent 90.72\% of the total apps including ML libraries in their codebase (Table~\ref{tab:dataset}), with the rest accounting for obfuscated, encrypted or lazily downloaded models.
In total these account for 1,666 models -- 1436 (86.19\%) \texttt{TFLite}, 176 (10.56\%) \texttt{caffe}, 46 (2.76\%) \texttt{ncnn}, 5 (0.3\%) \texttt{TensorFlow} and 3 (0.18\%) \texttt{SNPE}.
\texttt{TFLite} is expectedly first in popularity, as the recommended solution from the OS provider for mobile ML inference. However, it is surprising to see \texttt{caffe} so widely used, since it has been long deprecated and replaced by \texttt{caffe2} in 2017 and now PyTorch Mobile. 

\noindent
\textbf{Observations:} \emph{These results illustrate a long latency between the \cready{state-of-the-art} frontier of ML frameworks and \cready{their adoption for in-the-wild deployment.}} 

\vspace{-0.2cm}
\subsection{Model categorisation}
\label{sec:models}

Here, we perform a quantitative analysis of DNN models and their respective apps and correlate them with metadata from the Google Play Store. Our aim is to categorise the most popular DNN-powered apps and characterise their usage.

\begin{figure}[t]
\centering
\vspace{-0.2cm}
\includegraphics[width=0.44\textwidth]{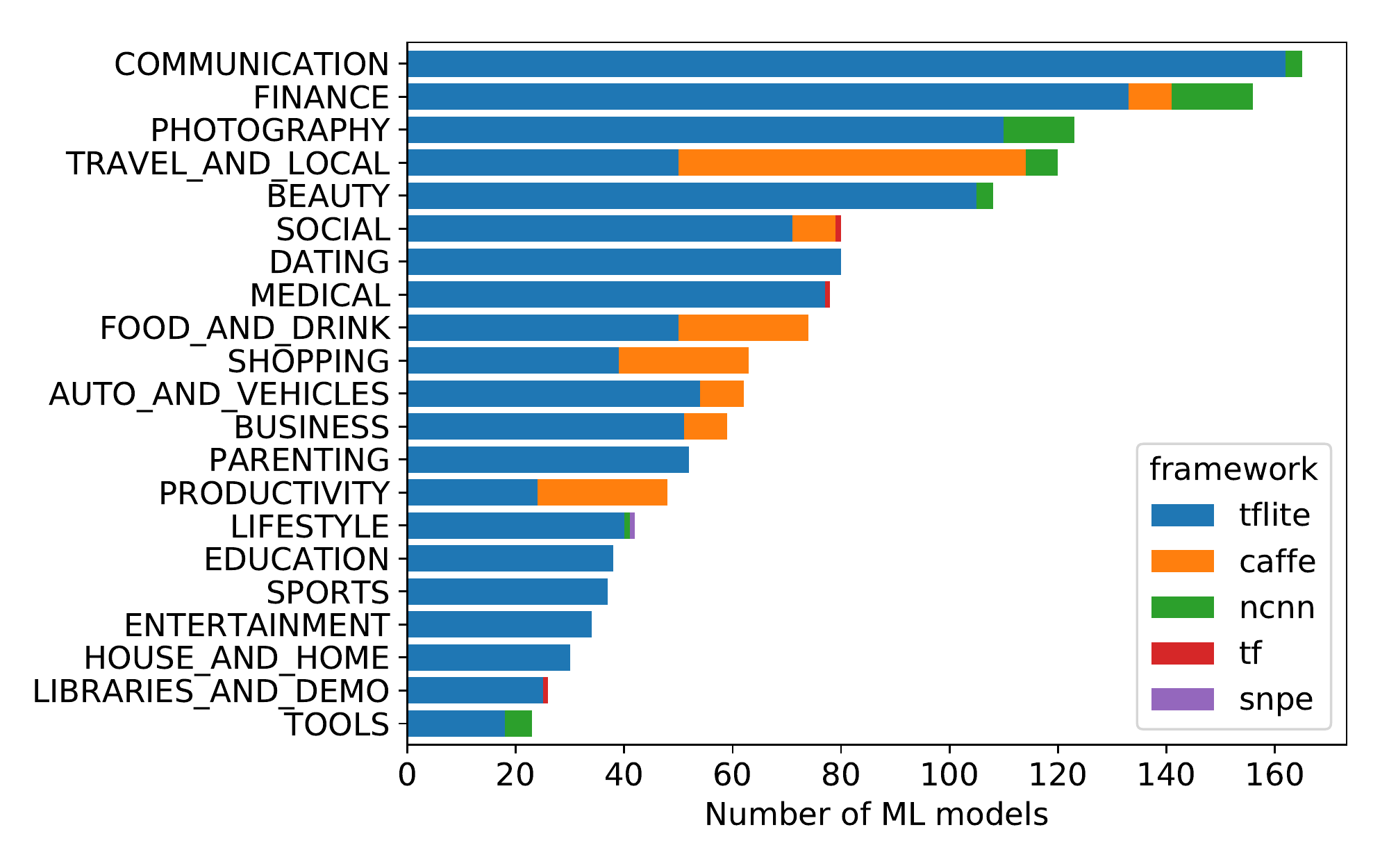}
\vspace{-0.4cm}
\caption{Number of models \tool{} successfully extracted and executed per framework and \cready{Google Play} category. Categories with less than 20 models are excluded.}
\vspace{-0.4cm}
\label{fig:categories}
\end{figure}

Fig.~\ref{fig:categories} \cready{shows the number of ML models per framework and Google Play category.}
We observe that the top DNN-powered apps belong to ``communication'' and ``finance'' tools with several DNNs for face and object detection (e.g. for detecting a card or ID to make transactions in the latter case). These are followed by more traditionally DNN-backed categories, such as ``photography'' and ``beauty'', which typically contain DNN-based filters to enhance photos. Potentially less expected categories 
include ``food and drink'', ``dating'' and ``parenting''. By manually examining these models, we found anecdotal examples of apps within these categories using DNNs to detect or recognise objects (e.g. a bottle of wine or a face), for recommendation systems (e.g. partner matching, advertising and food recipe recommendation) and even for baby monitoring.

To dig deeper into the purpose of each AI model, we manually looked into the naming, input/output dimensions and layer types of the encountered DNN models in order to characterise their usage. This labour intensive job was done across three ML researchers with a majority vote on the results.
We were able to identify the usage of $1,531$ models, accounting for $91.9\%$ of all models, with around $67\%$ having names which hint either the model, task at hand or both (e.g. ``hair\_segmentation\_mobilenet.tflite''). 
Our characterisation shows that the most popular task for deploying Deep Learning is computer vision ($>89\%$ of all models), followed by NLP (17 models) and audio (15 models). Last, we found traces of DNN models (4 models) utilising sensor data, such as accelerometer, gyroscope, etc. 
Two anecdotal use-cases for sensor ML are horse movement tracking and car crash detection in insurance apps.
Task-specific results are shown in Table~\ref{tab:dnn_tasks}, where it can be seen that most vision models were targeted at object, face and contour detection, most audio tasks at ambient sound recognition, most NLP tasks at text-completion and sensor tasks at movement tracking.

\noindent
\textbf{Observations:} \textit{Vision models seem to be the most prevalent, with a focus on object and face detection and text recognition and used mostly across communication, photography and beauty apps.}

\vspace{-0.2cm}
\subsection{Model uniqueness characterisation.}
Diving deeper into the models distributed amongst the most popular applications, we found that not all models are bespoke or unique.
Overall, we witness DNN models spread across different application categories, with a significant portion of these being off-the-shelf models without customisation. 
In fact, after checking for unique checksums on these models \cready{and respective weights}\footnote{\cready{Most apps distribute the model weights in their \texttt{apk}, either in a single file, along with the DNN graph, or in separate files (e.g. \texttt{caffe}). In either case, we perform an \texttt{md5} checksum on both the model and weights.}}, we find that only 318 models (19.1\% of the models as shown in Table~\ref{tab:dnn_tasks}) are unique.
For the most prevalent vision task, i.e., object detection, FSSD~\cite{li2017fssd} seems to be the most popular model. 
We found such occurrences even within popular Google apps (e.g. ``Gallery Go'' and ``Arts \& Culture'').
For face detection the Blazeface~\cite{bazarevsky2019blazeface} is another very popular model.
Spanning across tasks, MobileNet~\cite{howard2017mobilenets} seems to be the most popular architecture with variants (e.g. FSSD) being used to other vision tasks including semantic segmentation, pose estimation or classification.
Last, we encounter multiple occurences of models tackling a common task, e.g. recognise information from credit cards \cite{paycards_ios}, such as names and dates. %

\begin{table}[t]
\begin{subtable}[t]{.49\columnwidth}
    \centering
    \footnotesize
    \begin{tabular}{lr}
    \hline
    Task & Models \\ \hline
    \multicolumn{2}{c}{\scriptsize{\textbf{Vision}} (1495 models)} \\
    object detection        & 788 ($52.7\%$) \\
    face detection          & 197 ($13.2\%$) \\
    contour detection       & 192 ($12.8\%$) \\
    text recognition        & 185 ($12.4\%$) \\
    augmented reality       & 51  ($3.4\%$) \\
    semantic segmentation   & 14  ($0.9\%$) \\
    object recognition      & 14  ($0.9\%$) \\
    pose estimation         & 8   ($0.5\%$) \\
    photo beauty            & 8   ($0.5\%$) \\
    image classification    & 7   ($0.4\%$) \\
    nudity detection        & 5   ($0.3\%$) \\
    other                   & 26  ($1.7\%$) \\
    \hline
    \end{tabular}
\end{subtable}
\hfill
\begin{subtable}[t]{.49\columnwidth}
    \footnotesize
    \begin{tabular}{lr}
    \hline
    Task & Models \\ \hline
    \multicolumn{2}{c}{\scriptsize{\textbf{NLP}} (17 models)} \\
    auto-complete           & 9 ($52.9\%$) \\
    sentiment prediction    & 4 ($23.5\%$) \\
    content filter          & 2 ($11.8\%$) \\
    text classification     & 1 ($5.9\%$) \\
    translation             & 1 ($5.9\%$) \\
    \hline
    \multicolumn{2}{c}{\scriptsize{\textbf{Audio}} (15 models)} \\
    sound recognition     & 12 ($80.0\%$) \\
    speech recognition    & 2  ($13.3\%$) \\
    keyword detection     & 1  ($6.7\%$) \\
    \hline
    \multicolumn{2}{c}{\scriptsize{\textbf{Sensor}} (4 models)} \\
    movement tracking    & 3 ($75.0\%$) \\
    crash detection      & 1 ($25.0\%$) \\
    \hline
    \end{tabular}
\end{subtable}
\vspace{0.1cm}
\caption{DNN task classification.}
\vspace{-0.9cm}
\label{tab:dnn_tasks}
\end{table}

\noindent
\textbf{Model fine-tuning.} 
Taking this analysis one step further, we perform a checksum-based analysis at finer-granularity (layer-level) to see to what degree to developers train their own models from scratch or fine-tune the last layers through \textit{transfer learning} \cite{pan2009survey}. The intuition is that the first layers of the network are typically extracting low-level features (e.g. edges, shapes, etc. for vision tasks) that are shared between similar tasks and only deeper in the DNN do the task-specific and semantically relevant features get extracted.
Results from our analysis show that, excluding  duplicate models, 9.02\% of the remaining models share  at least 20\% of the weights  with at least one other model. 
In fact, $4.2\%$  of the models only differ in up to three layers, indicating that some developers only fine-tune small portions of the network, resulting in a significantly smaller training footprint and exploiting  transfer learning from other (typically off-the-shelf) networks.
Moreover, we checked for traces of online fine-tuning done on device (e.g. through \texttt{TFLiteTransferConverter} \cite{tflite_personalisation})
and found none, indicating that on-device fine-tuning is not yet widely exploited in the wild due to the significant computation requirements and the limited availability of labelled high-quality on-device datasets.

\noindent
\textbf{Observations.} \textit{Based on this type of evidence, we deduce that it is common for developers to leverage a pre-trained model that is widely available and pay the significantly smaller cost of training offline only a subset of the last DNN layers. While online on-device training is a prominent future avenue, be it through fine-tuning or federated learning, current support in mobile frameworks is limited and so are such deployments.}

\vspace{-0.2cm}
\subsection{Temporal analysis across snapshots}
\label{sec:temporal}

\begin{figure}[t]
\centering
\includegraphics[width=0.95\columnwidth]{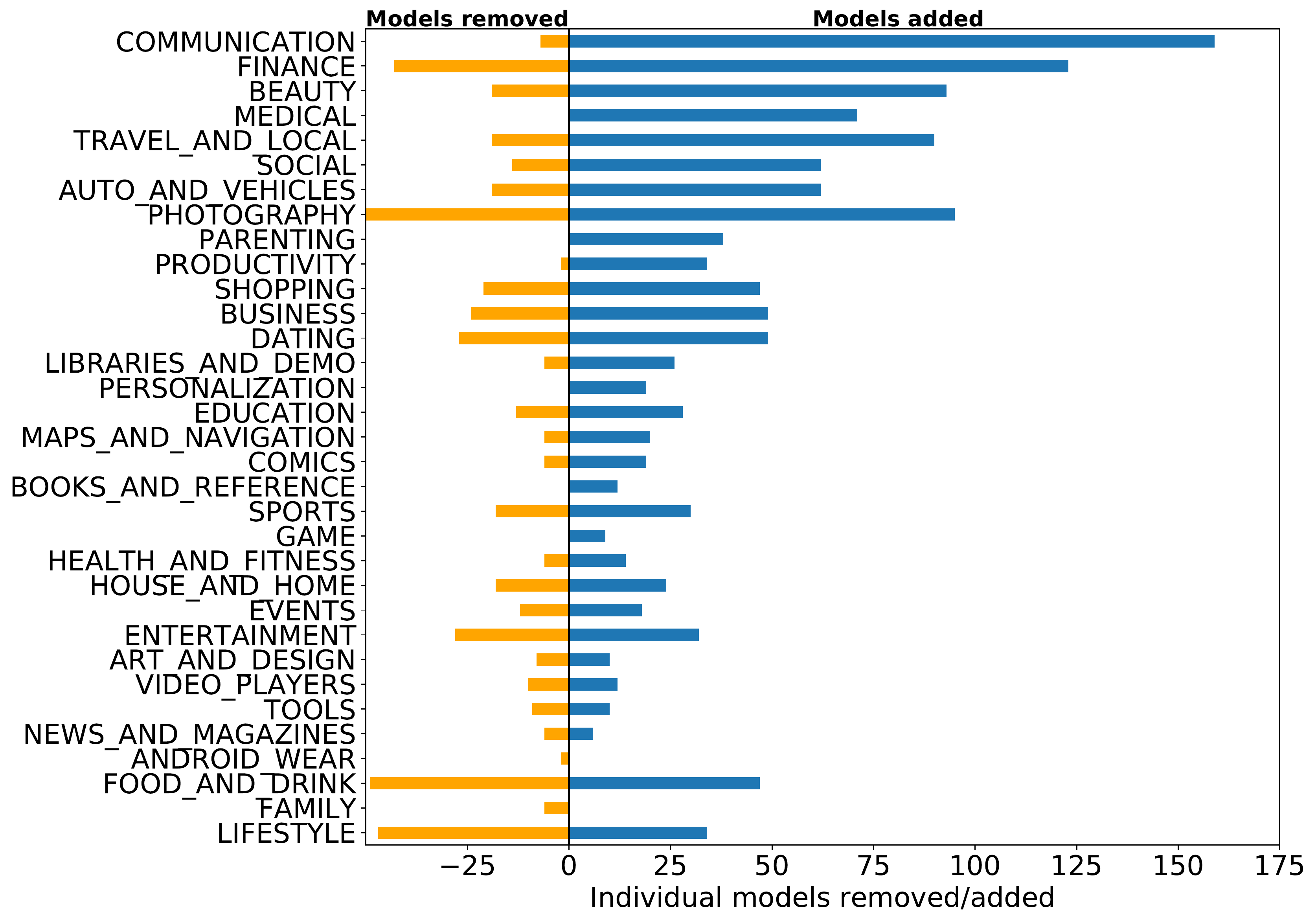}
\vspace{-0.4cm}
\caption{Individual models removed/added \cready{between two snapshots taken one year apart.}}
\vspace{-0.4cm}
\label{fig:temporal_models_added_removed}
\end{figure}

As aforementioned, we took two distinct snapshots of the most popular apps in the Google Play Store 12 months apart from each other.
In this part of our analysis, we compare and contrast these two snapshots in terms of app popularity and in-the-wild DNN deployment and draw conclusions about the trajectory of ML penetration in smartphones nowadays.
What is unique about our dataset is that we happened to measure DNN-deployment across the COVID-19 pandemic, which had a crucial impact on human activity during the course of 2020/2021. For this reason, we also compare our temporal analysis with similar analyses done in the past \cite{xu2019first} to i)~identify potential biases of our dataset during these exceptional circumstances and ii)~to see how app popularity and, as an extension, DNN adoption, has been affected by these circumstances.

Results from our temporal analysis indicate a surging number of DNN models being deployed on the Android platform, essentially doubling in the course of 12 months. Specifically, our traced models went from $821$ to $1.6K$ for our latest snapshot (Table~\ref{tab:dataset}), with most additions belonging to vision tasks. 
\texttt{TFLite} remains the dominant mobile inference framework, going from $81.6\%$ to $86.1\%$ of the total models found ($2.15x$). 
The increase in models was less pronounced for \texttt{ncnn} ($1.18x$) and \texttt{caffe} ($1.69x$).
The latter is surprising given the fact it has been deprecated and newer frameworks have taken its place (\texttt{caffe2} and PyTorch Mobile).
Finally, we observe a drop in the \texttt{TF} ($0.56x$) adoption rate, which is expected given the increasing popularity of its mobile counterpart.

Next, we analyse the DNN models across snapshots per category of application to which they belong. Fig.~\ref{fig:temporal_models_added_removed} depicts the number of individual models that were removed/added across our snapshots, sorted by the difference between the two.
Interestingly, most additions of ML models happened for communication tools, taking the lead from ``photography'' applications, which was the top ML-powered category of 2020. This can potentially indicate that communication apps became more important due to the pandemic, and developer focus was diverted to this category. A similar trend could be witnessed for ``finance'' applications, where we observed many models aimed at the automated identification of people and their ID cards. Whilst this traditionally constituted a manual process done in person in financial institution (e.g. banks), the pandemic might have created a new need for ML models to fill. Last, apps related to ``health'' and ``medical'' purposes seem to have a surging deployment of DNN models. On the other side of the spectrum, ``lifestyle'', ``food \& drinks'' and ``Android Wear'' applications seem to be falling in terms of popularity, something that could be potentially attributed to the fact that people stay more at home.

Next, we integrate the results of previous analyses~\cite{xu2019first,sec_dnns_apps} to shape a more general trend for DNN adoption in the Android ecosystem. 
In \cite{xu2019first}, the authors report the total ML-backed apps going from $166$ in June 2018 to $211$ in September 2018. 
In \cite{sec_dnns_apps}, the authors traced $178$ ML-powered apps, somewhere between \cite{xu2019first} and June 2020 \footnote{The snapshot date is not reported, thus we consider it between \cite{xu2019first}, with which it compares, and the work's venue submission date.}. 
Last, for our trace, we report ML-powered apps going from $236$ to $377$ from February 2020 to April 2021.
From the previously reported figures, we witness a soaring trajectory of ML apps deployed in the wild, with the adoption rate of ML being accelerating.

\noindent
\textbf{Observations:} \textit{\cready{While there was a big reshuffling in the type of AI models deployed during the pandemic}, we observe a considerable \cready{general growth in the number of DNN models in AI-powered applications in the past 3 years (from 176 in 2018}~\cite{xu2019first} \cready{to 1,666 in April 2021}). These results demonstrate how the proliferation of mobile AI frameworks, the availability of pre-trained models and the constant improvement of mobile hardware have driven this growth and the need to keep up with this ever-increasing adoption.}

\begin{figure}[t]
    \centering
    \includegraphics[width=0.8\columnwidth]{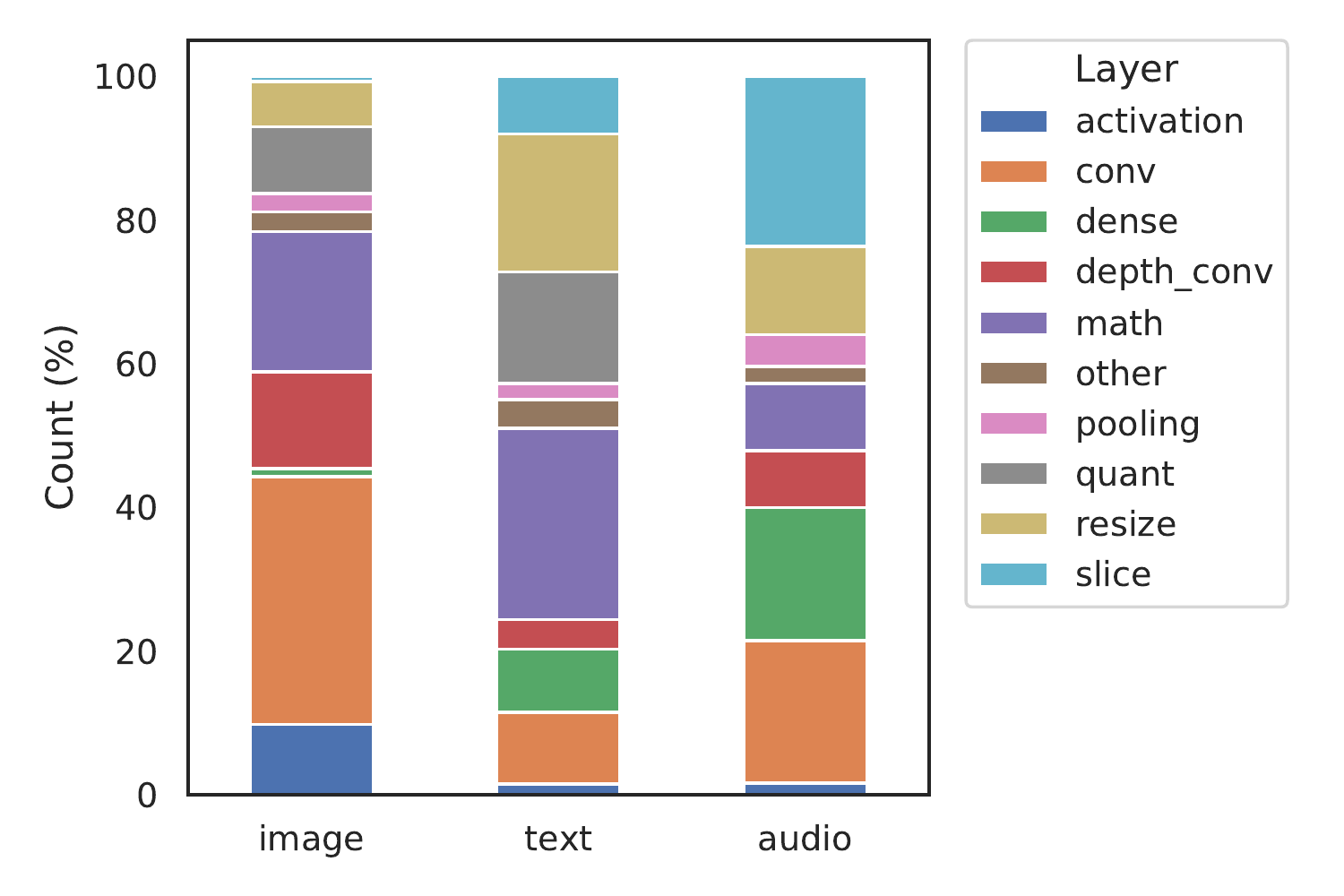}
    \vspace{-0.44cm}
    \caption{Model layer composition per input modality for \texttt{TFLite}, \texttt{NCNN} and \texttt{caffe}.}
    \vspace{-0.4cm}
    \label{fig:layers}
\end{figure}

\vspace{-0.2cm}
\subsection{Mobile DNNs layers and operations}
\label{sec:dnn_ops}

After having coarsely characterised the models based on their input modality, target task and app category, we take a finer-grained look into the models and analyse their structure in terms of the layers and operations they contain.

\noindent
\textbf{DNN layers and operation types.}
First, we go through the graph representing each DNN and trace the layer types they contain, grouping results per input modality. Results are shown in Fig.~\ref{fig:layers} for \texttt{TFLite}, \texttt{NCNN} and \texttt{Caffe}. We see \textit{convolution} layers being amongst the most popular layer types across modalities (34\%, 10\%, 20\% for image, text and audio, respectively). Originally applied in visual tasks, their usage nowadays spreads across recommender systems, natural language processing and time-series analysis. Variants such as \textit{depthwise-separable convolutions} (\texttt{depth\_conv}) \cite{howard2017mobilenets} are computationally less heavy and are aimed for mobile deployments. \textit{Dense} (or \textit{linear}) layers are fully-connected layers that are typically found in the output of classification tasks, or in the implementation of RNNs. Majority of these layers are found in audio (19\%) and text (9\%) models. \textit{Activations} essentially impose non-linearity in DNNs, and can be fused with the previous layer in terms of implementation. Thus, the existence of such operations as distinct layers is framework dependent. Last, ``helper'' layers such as \textit{math}, \textit{quant}, \textit{resize} and \textit{slice} operations are performing math or matrix representation operations and can be found across modalities. 

\begin{figure}[t]
    \centering
    \includegraphics[width=0.9\columnwidth]{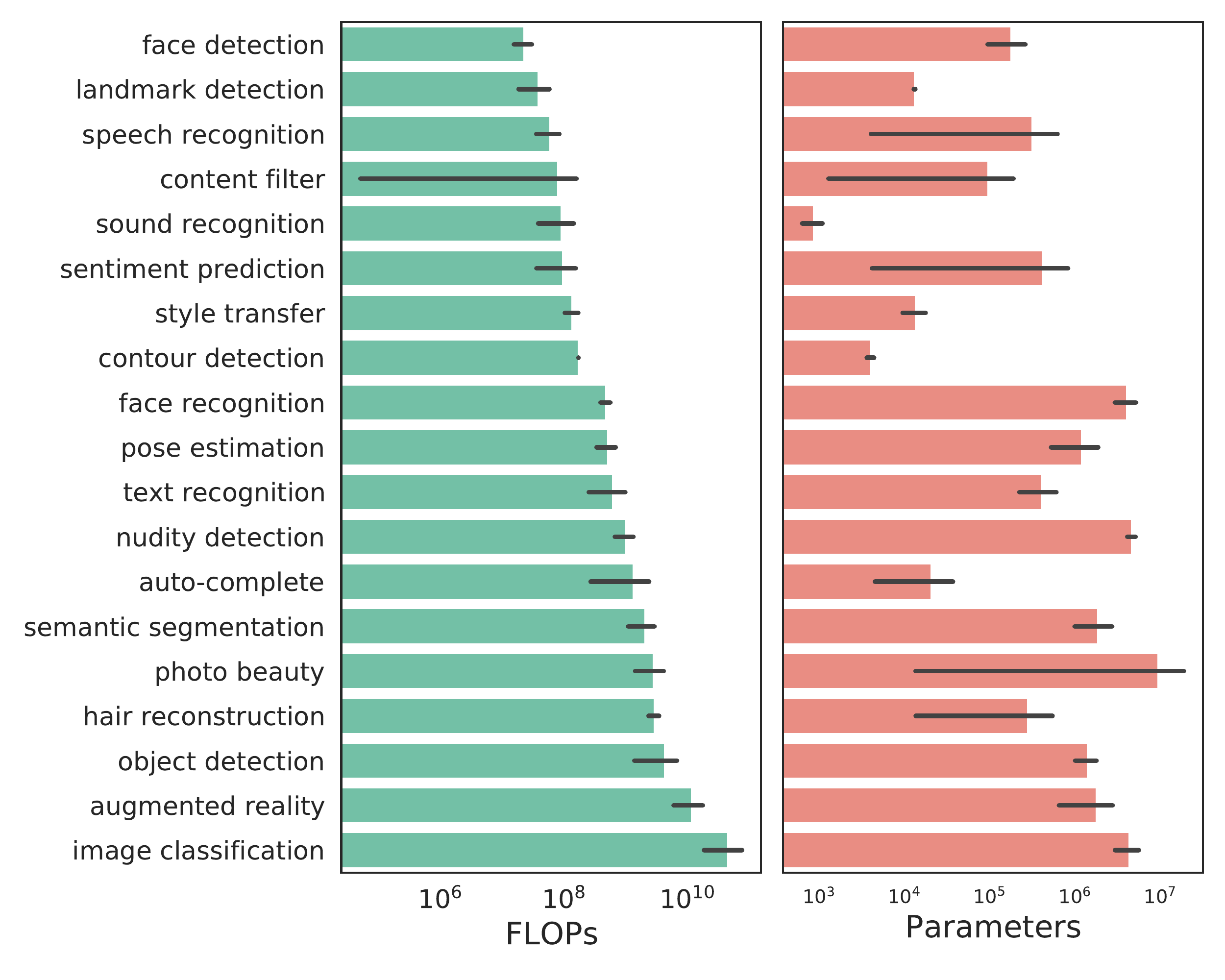}
    \vspace{-0.4cm}
    \caption{FLOPs and parameters per DNN task.}
    \label{fig:flops-new}
    \vspace{-0.5cm}
\end{figure}

\begin{figure}[b]
    \vspace{-0.5cm}
    \centering
    \begin{subfigure}{\columnwidth}
        \centering
        \includegraphics[width=\columnwidth]{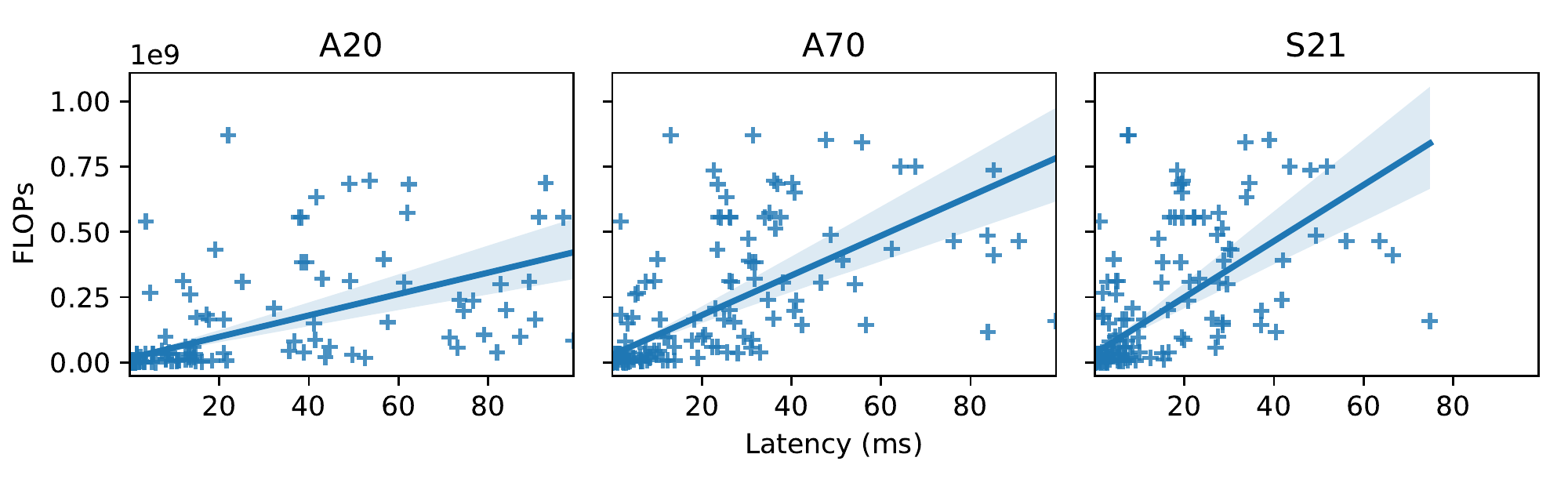}
        \vspace{-.3cm}
    \end{subfigure}
    \begin{subfigure}{\columnwidth}
        \centering
        \includegraphics[width=\columnwidth,clip,trim={0 0 0 4.83cm}]{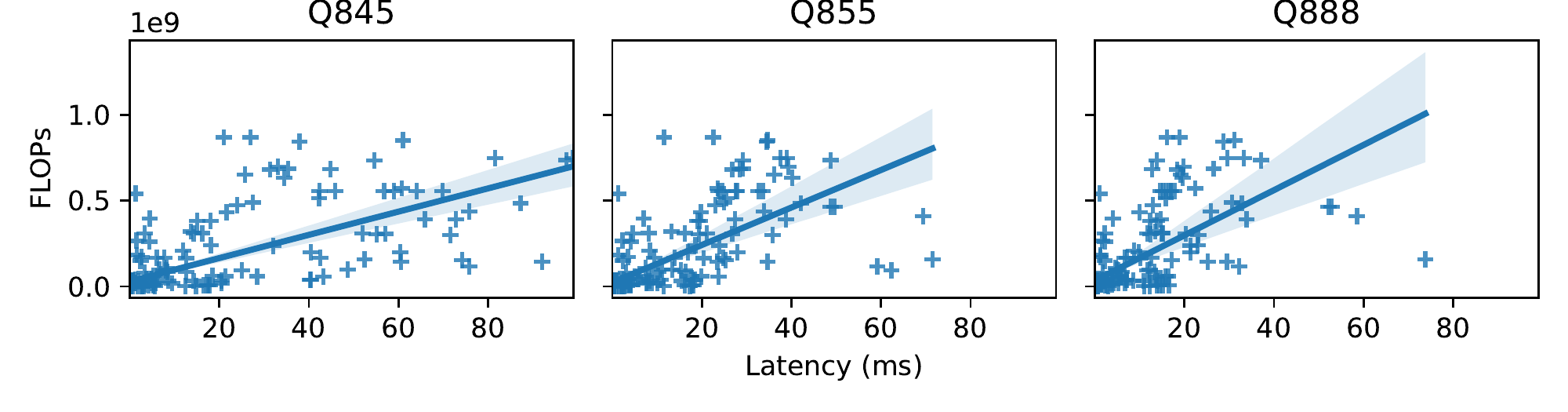}
    \end{subfigure}
    \vspace{-0.4cm}
    \caption{\cready{Observed relationship between latency and FLOPs across six different devices.}}
    \label{fig:latency-flops-scatter}
\end{figure}

\noindent
\textbf{DNN \#operations and \#parameters.} Next, we estimate the number of operations (in FLOPs) and parameters that each model contains by going through the graph in a trace-based manner. \cready{Concretely, we generate a random input with the DNN-specified input dimensions and perform a DNN inference. During the forward propagation step, we measure analytically the amount of operations being performed per layer (dependent on the kind of layer) and the number of trainable parameters associated with it.}
Fig.~\ref{fig:flops-new} shows the result of this analysis per DNN task. We see that among the traced models, on average the heaviest deployed vision models belong to classification, hair reconstruction, segmentation and beauty tasks. For NLP the heaviest tasks belong to text auto-completion whereas for audio the heaviest deployed task is sound recognition.
At this point, we note that these numbers only refer to the traced deployed models and \cready{do} not represent a generic commentary on the overhead of models per task. In fact, in many cases it is the opposite if we only take into consideration the task (e.g. classification vs. segmentation or speech vs. sound recognition). Also, we note that the number of models found for each task category varies significantly.

\noindent
\textbf{Observations:} \textit{We find that convolutions dominate the mobile DNN landscape due \cready{to their wide use in vision models}, as well as  the fact that they can map well on mobile hardware for efficient execution, compared to e.g. recurrent layers \cite{10.1109/MICRO.2018.00022}.
While depth-wise convolutions can significantly improve performance, their deployments are scarcer as they can impact the quality of the model. Furthermore, we find that there is huge variance in terms of FLOPs and parameters (four orders of magnitude) in the traced models. This might be attributed to the granularity of the task corresponding to a single inference. For example, in image recognition the input is typically an RGB image while in next-word prediction the input can be a couple of words.}

\section{Runtime analysis of Mobile DNNs}
\label{sec:benchmarking}
Up until now, we have focused our efforts on analysing the DNN models in an offline manner.
In this section, we turn to on-device benchmarking and report on performance and energy when running the encountered models across the devices presented in Table~\ref{tab:specs}. This analysis provides important insights about how real-world AI applications are performing on a heterogeneous set of devices, thus answering \textbf{RQ\#2}.

\subsection{On-device DNN latency}
\label{sec:on-device-lat}

\begin{figure}
\centering
\includegraphics[width=0.8\columnwidth]{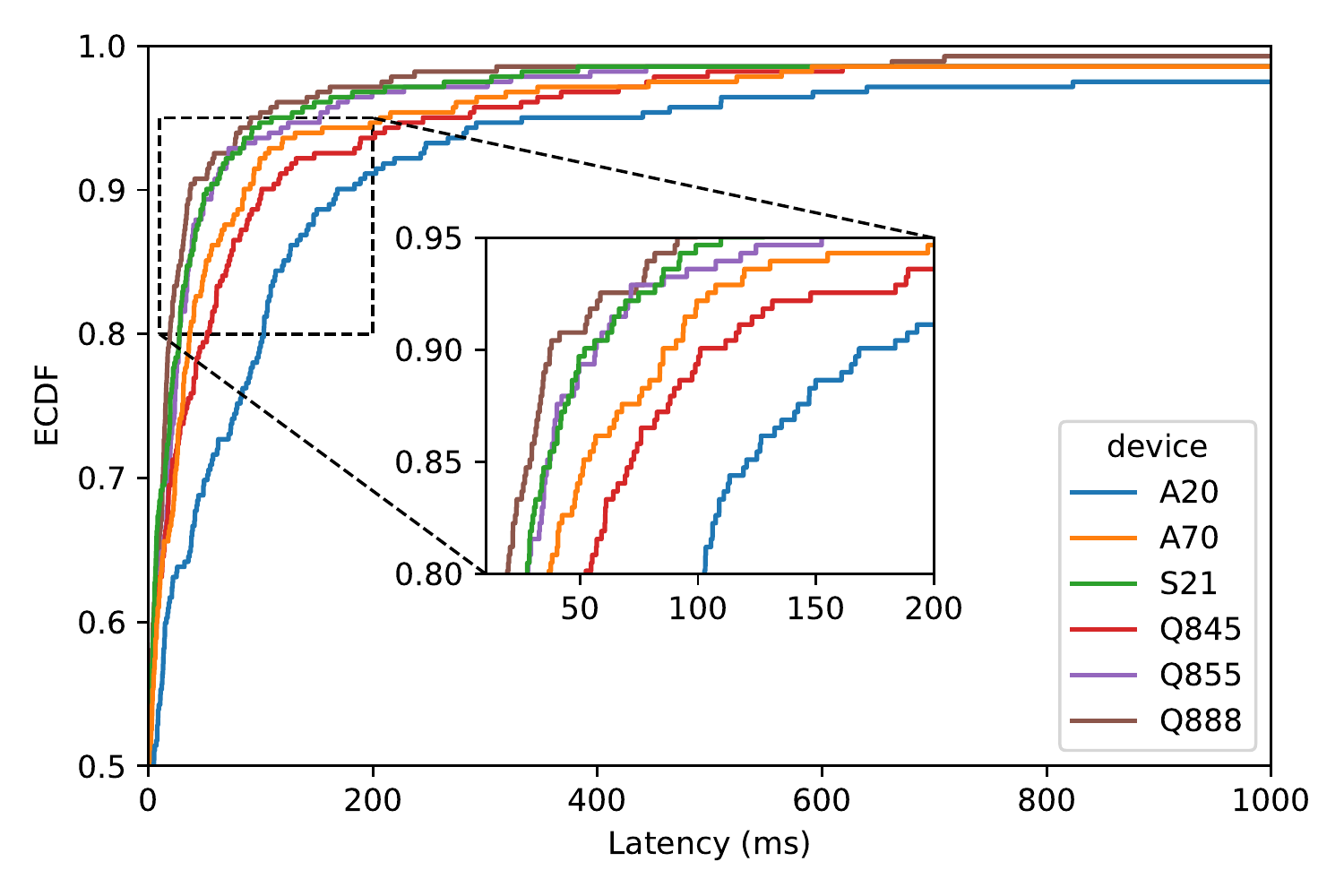}
\vspace{-0.4cm}
\caption{Latency per device ECDF.}
\vspace{-0.4cm}
\label{fig:latency-ecdf}
\end{figure}

\begin{figure*}[t]
    \centering
    \begin{subfigure}{0.32\textwidth}
        \includegraphics[width=\textwidth]{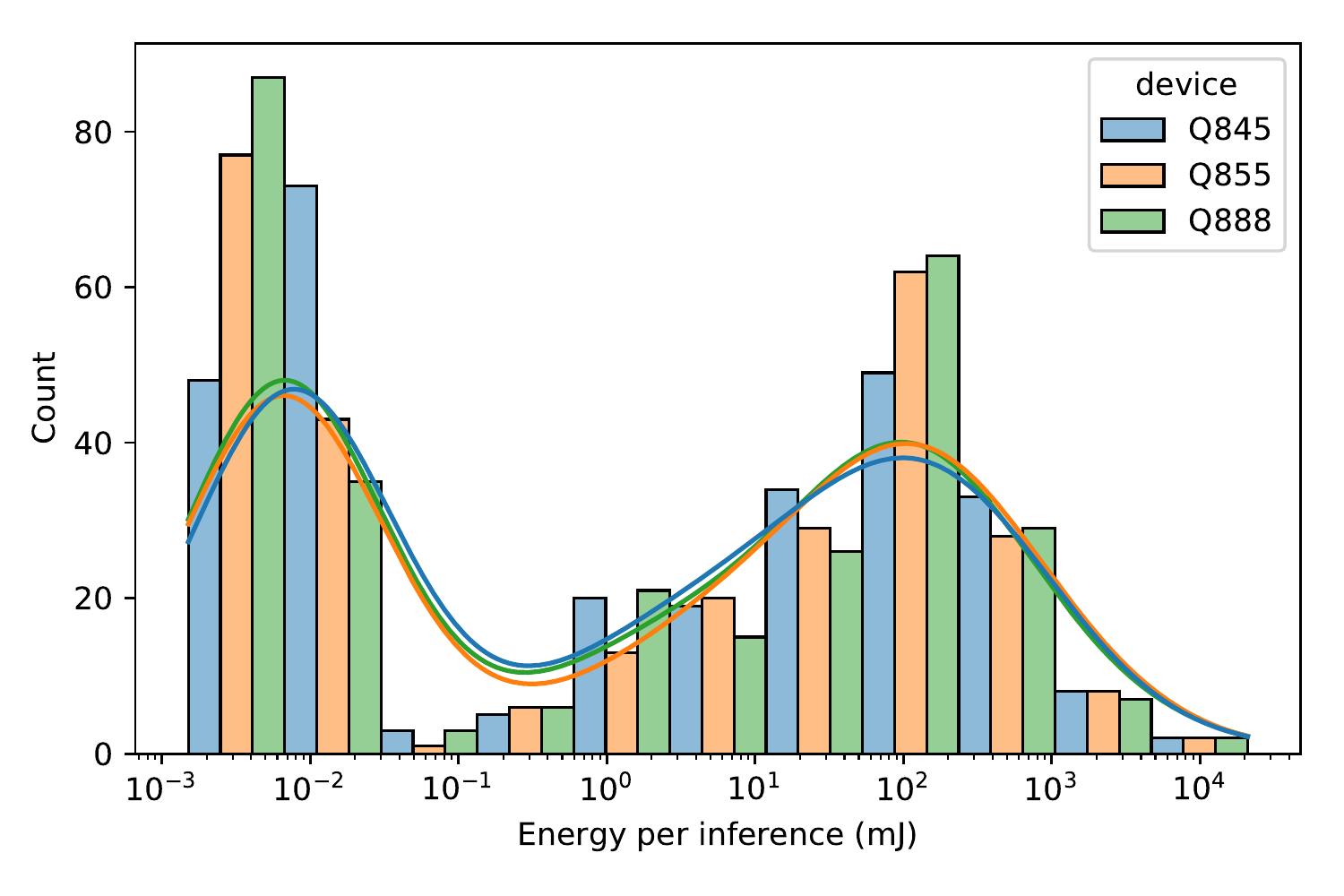}
        \vspace{-0.7cm}
        \caption{Inference energy}
        \label{fig:inference_energy}
    \end{subfigure}
    \hfill
    \begin{subfigure}{0.32\textwidth}
        \includegraphics[width=\linewidth]{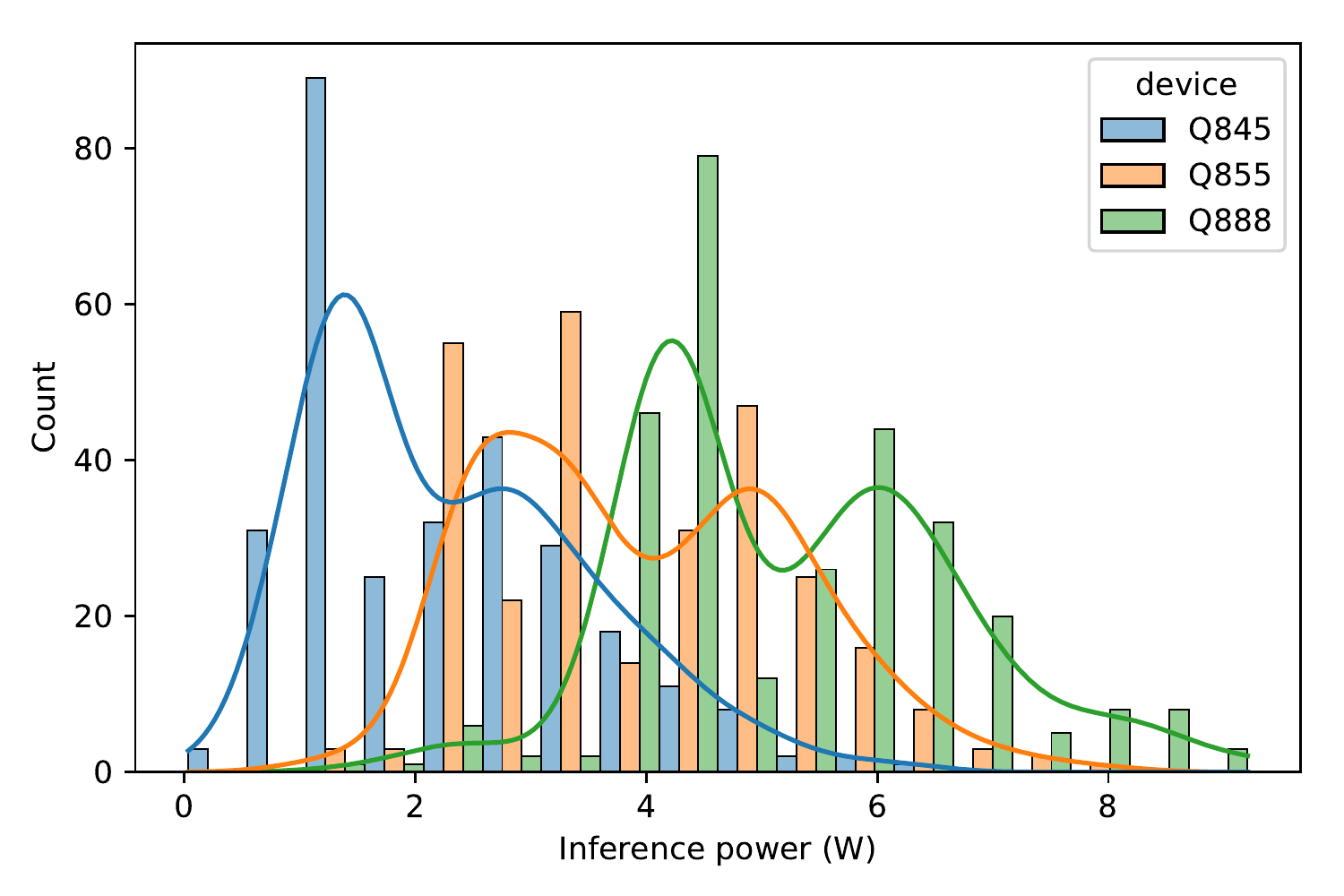}
        \vspace{-0.7cm}
        \caption{Inference power}
        \label{fig:inference_power}
    \end{subfigure}
    \hfill
    \begin{subfigure}{0.32\textwidth}
        \includegraphics[width=\linewidth]{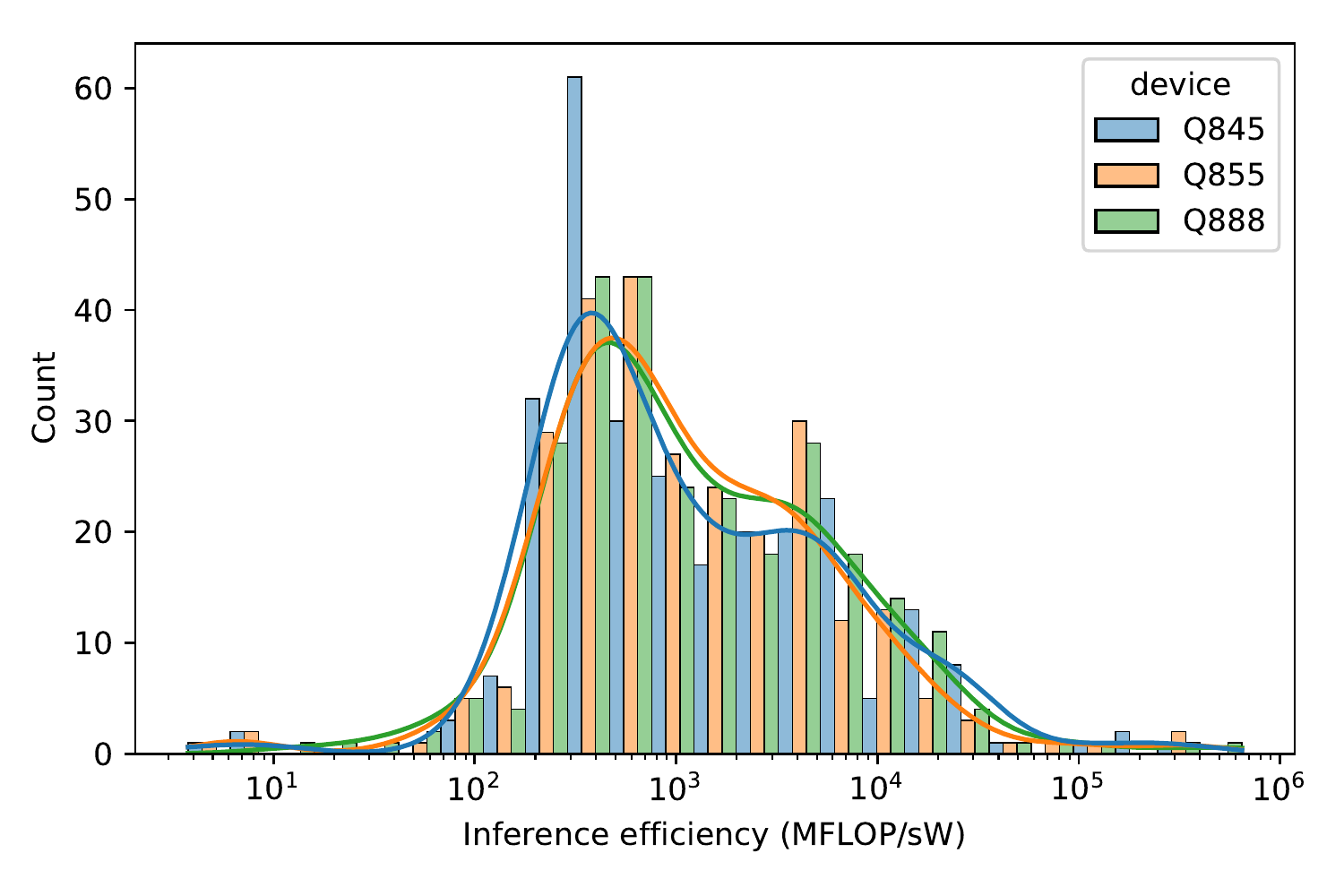}
        \vspace{-0.7cm}
        \caption{Inference efficiency}
        \label{fig:inference_efficiency}
    \end{subfigure}
    \vspace{-0.3cm}
    \caption{Distributions of inference energy, power and efficiency of the collected models when run across 3 generations of Qualcomm SoCs. The lines represent kernel density estimations.}
    \vspace{-0.4cm}
    \label{fig:energy_power_hist}
\end{figure*}

Prior work \cite{almeida2019embench,ai_benchmark_2019} has shown that FLOPs is not necessarily a good proxy for estimating a model's on-device performance. Reasons for such discrepancies include the underutilisation of hardware due to e.g. memory-bound operations, thermal throttling due to continuous inference or even due to scheduling on cores of different dynamics due to energy-saving scheduler policies on Heterogeneous Multi-Processors \cite{kim2017enhancing}.
To further corroborate this fact, \cready{in Fig.~\ref{fig:latency-flops-scatter} we depict the FLOPs and actual measured inference latency across devices for different models.
Our analysis on real-world models on different devices reinforces this non-linear (line-fit) relationship as it not only varies for different model architectures, but also differs from one device to another.}

\cready{To investigate this further}, in  Fig.~\ref{fig:latency-ecdf} we show the ECDF of model runtime across all available devices. From the graph it is evident that the computing gap between  a \textit{low-end} device (A20) and a \textit{mid-tier} device (A70) is considerably larger than the difference of \textit{mid-tier} to \textit{high-end} (S21). Specifically, \textit{low-end} and \textit{mid-tier} devices (A20 and A70) are $3.4\times$ and $1.51\times$ slower compared to S21. 
Across generations of high-end SoCs of the same manufacturer (Q845, Q855, Q888), we see incremental performance gains (i.e., average latency of $76$, $58$ and $35$ ms), but noticeable, to the point that a next-gen mid-tier phone may perform better than the high-end SoC of a prior generation, despite claims about significant boosts in AI acceleration between generations. Last, we want to mention that for the two devices that integrate the same SoC (Q888 and S21), the open-deck design of the development board along with the vanilla variant of the OS leads to incrementally better results and faster inference overall. Heat dissipation of the open design, cross-manufacturer configurations and low-level configuration of the Android Scheduler can all be contributing factors.

\noindent
\textbf{Observations:} \textit{We observe a wide variability of inference latency across devices even for models that have similar FLOP counts, which reaffirms the need for on-device benchmarking. Devices of different tiers and generations offer variable dynamics, with the lower-tier falling significantly behind in performance. Even devices integrating the same SoC can offer variable performance due to vendor-specific configurations, the installed apps and drivers or even due to different thermal characteristics. Therefore, given this heterogeneity, it is hard for developers to accurately predict the users' experience without testing their models on a large sample of devices.}

\subsection{Energy consumption}

In mobile settings, one cannot simply optimise for performance without taking energy consumption into consideration. While smartphone capabilities are growing larger every year, the same developments have not been witnessed in battery technology. Therefore, quantifying the cost of being smart in terms of energy is an important component in the mobile world. In this section, we report on the energy, power and efficiency of doing inference on device, across frameworks for the three Snapdragon boards representing different generations of devices.

\subsubsection{Energy and power consumption per device}

Fig.~\ref{fig:inference_energy} shows the distribution of models with respect to the energy required per inference across our three devices.
Expectedly, we see from the kernel density function lines that all three devices follow a similar trajectory, indicating that a similar amount of energy is required for similar workloads regardless of the device.
On the other hand, this is not the case in terms of power consumption (Fig.~\ref{fig:inference_power}), where we can see newer generations of devices consistently drawing more power to run models.
This is a direct implication of the fact that newer generations of devices can execute models faster, as shown in Fig.~\ref{fig:latency-ecdf}, while energy required remains similar.

Following these observations, we decided to calculate inference efficiency per each model by calculating how many floating-point operations can be executed in one second per one Watt\footnote{Effectively the same as calculating FLOPs per Joule.}.
As can be seen in Fig.~\ref{fig:inference_efficiency}, trends in efficiency stay mostly the same across different devices, following energy consumption, but unlike energy we can see a minor improvement of the newer devices over Q845 in the middle of the distribution, suggesting that relatively more models can run more efficiently (median efficiency of 730, 765 and 873 MFLOP/sW, after removing outliers) on the newer hardware.

\subsubsection{Use-case driven energy consumption}
\label{sec:energy-scenarios}
Up to here, we have seen performance and energy consumption for single inferences. However, the quanta of data associated with each inference may vary considerably between tasks or modalities as noted before in Sec.~\ref{sec:dnn_ops}. 
Thus, we dive deeper into three selected tasks representative of each modality, namely i)~\textit{sound recognition} for audio, ii)~\textit{auto-completion} for text and iii)~\textit{semantic segmentation} for vision.

\begin{table}[]
    \centering
    \small
    \begin{tabular}{@{}lcccc@{}}
        \toprule
         \multirow{2}{*}{\centering \textbf{Use-case}} & \multicolumn{4}{c}{\textbf{Battery discharge (mAh)}} \\
          \cmidrule(l){2-5}
         & Avg. & Median & Min & Max \\ \midrule
         \multicolumn{5}{l}{\textbf{Q845}} \\ %
         Sound R. & 0.6350$\pm$2.0226 & 0.0652 & 0.0351 & 2.5277 \\
         Typing   & 0.0752$\pm$0.1637 & 0.0292 & 0.0245 & 0.1993 \\
         Segm.    & 1221.7$\pm$2761.0 & 619.62 & 271.93 & 3835.2 \\ %
         \hline
         \multicolumn{5}{l}{\textbf{Q855}} \\ %
         Sound R. & 1.0311$\pm$3.3438 & 0.1821 & 0.0262 & 5.0327 \\
         Typing   & 0.1192$\pm$0.2835 & 0.0387 & 0.0279 & 0.3404 \\
         Segm.    & 1133.4$\pm$2468.1 & 489.10 & 262.85 & 3239.7 \\ %
         \hline
         \multicolumn{5}{l}{\textbf{Q888}} \\ %
         Sound R. & 0.7950$\pm$2.8060 & 0.1009 & 0.0316 & 4.4132 \\
         Typing   & 0.1001$\pm$0.2484 & 0.0315 & 0.0300 & 0.3403 \\
         Segm.    & 1062.7$\pm$2416.6 & 455.71 & 272.44 & 3290.8 \\ \bottomrule
    \end{tabular}
    \vspace{0.1cm}
    \caption{\cready{Scenario-driven energy consumption for three devices and use-cases in audio, text and vision.}}
    \vspace{-0.7cm}
    \label{tab:energy-usecases}
\end{table}

We make certain realistic assumptions on the data sizes, granularity input and frequency of results and then assess all relevant models belonging to this category.
Specifically, for \textit{speech recognition}, we assumed each model is run in order to recognize 1 hour of audio input.
To derive how long a model would need to be run, we manually investigated the models and assumed the most likely amount of audio input per inference considering the model's input dimension and common practices in speech ML~\cite{chan2015listen,Pratap2020scaling,mehrotra2021nasbenchasr}.
For \textit{text auto-completion} we assumed each model is run once per new word typed by a user, and further assumed a workload of 275 words, derived from WhatsApp's statistics about average daily number and length of messages~\cite{what-users, what-messages, what-words}.
Last, for \textit{semantic segmentation}, we assumed each model is used to segment a human at 15 FPS during a 1-hour-long video call in order to apply background effects, we further assumed that the model processes one frame per inference which is the usual approach~\cite{Long2015fcn,Zhao2018icn,chen2020fasterseg}.

Results \cready{across the development boards} are depicted in Table \ref{tab:energy-usecases} and indicate that different tasks and use cases result in very different impact on the battery life. 
On the high-end of energy consumption, we see that one hour of segmentation can result in a significant \cready {average reduction of 26.6\% to 30.54\% }of a common 4000mAh battery capacity (e.g.~A20 and S21).
\cready{Moreover, the most energy hungry segmentation models can almost deplete the full battery capacity within an hour, with a 80.9\% to 95.9\% reduction.}
On the other end, models like auto-completion are ubiquitous across messaging apps and deliver both in terms performance and efficiency, allowing their frequent use without a significant impact on battery. 

\noindent
\textbf{Observations.} \textit{Energy consumption is a major component in mobile, and intelligence comes at a  cost to battery life.
Unlike latency, which is visibly improved with new generations of devices, energy consumption seems to be predominantly dependent on the model architecture.
Even though newer hardware might improve in power-efficiency, differences are much less pronounced compared to performance improvements, which are even less observable across different model architectures. This suggests that it is the AI developers who can optimise battery life the most, unlike plain latency which can be improved at multiple levels, including manufacturers.
}

\section{Available Optimisations}
\label{sec:optimisations}
After examining how real-world DNNs run on a heterogenous set of devices, we now look into \textbf{RQ\#3} by means of DNN-specific as well as system-level optimisations aiming to improve inference and deployment performance. 

\begin{figure}[t]
    \centering
    \includegraphics[width=.85\columnwidth]{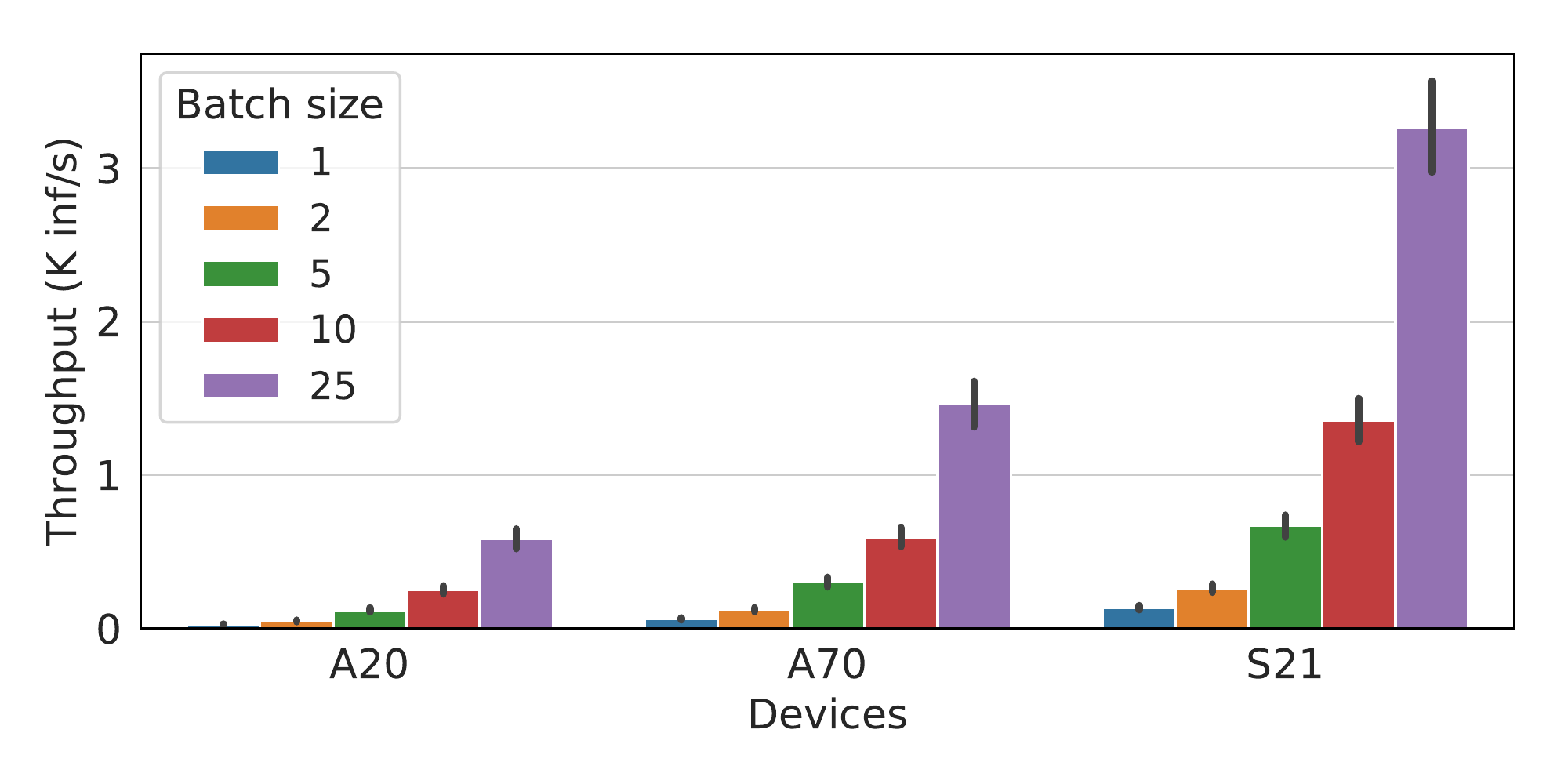}
    \vspace{-0.6cm}
    \caption{Inference throughput vs. batch size.}
    \vspace{-0.4cm}
\label{fig:batch}
\end{figure}

\subsection{Model-level Optimisations}

In this section, we focus on the adoption of three model-level optimisations, namely i)~\textit{weight clustering}, ii)~\textit{pruning} and iii)~\textit{quantisation}, for the identified \texttt{TFLite} models.

\noindent
\textbf{Clustering:}
\label{sec:clustering}
Clustering refers to the technique of reducing the number of distinct weight values by representing them through their clusters' centroids \cite{han2016deep}.
We identify clusters of shared weights by searching for layers with a \textit{``cluster\_''} prefix on \texttt{TFLite} models.
Despite the advertised potential for significant model size reductions \cite{tflite_clustering}, we report that none of the models in-the-wild seem to use weight clustering. This may be a result of either accuracy drops or the fact that the current clustering implementation does not reduce runtime memory and targets model compression only~\cite{tflite_clustering}.

\noindent
\textbf{Pruning:}
\label{sec:pruning}
Pruning refers to the technique of zero-ing out specific weights/channels of the network that have minimal impact on the output, due to representational redundancy in DNNs. 
Weight pruning can be detected during training by searching for layers with a \textit{``prune\_''} prefix for \texttt{TFLite} models. Nonetheless this prefix is often removed for inference~\cite{pruning}. We report that we did not find any occurrence of such layers either. 
While this approach has the potential to skip the zero weight computations during inference, the current implementation benefits only from increased sparsity \cite{tflite_pruning}
which, like clustering, results only in model compressibility.
To find if there is the potential of adopting magnitude-based weight pruning, we measured the weight sparsity for the tracked \texttt{TFLite} models. We find that, overall, 3.15\% of weights are near zero (within $\pm10^{-9}$), which might show limited prospects for weight magnitude-based pruning.

\begin{figure}[t]
    \centering
    \includegraphics[width=.85\columnwidth]{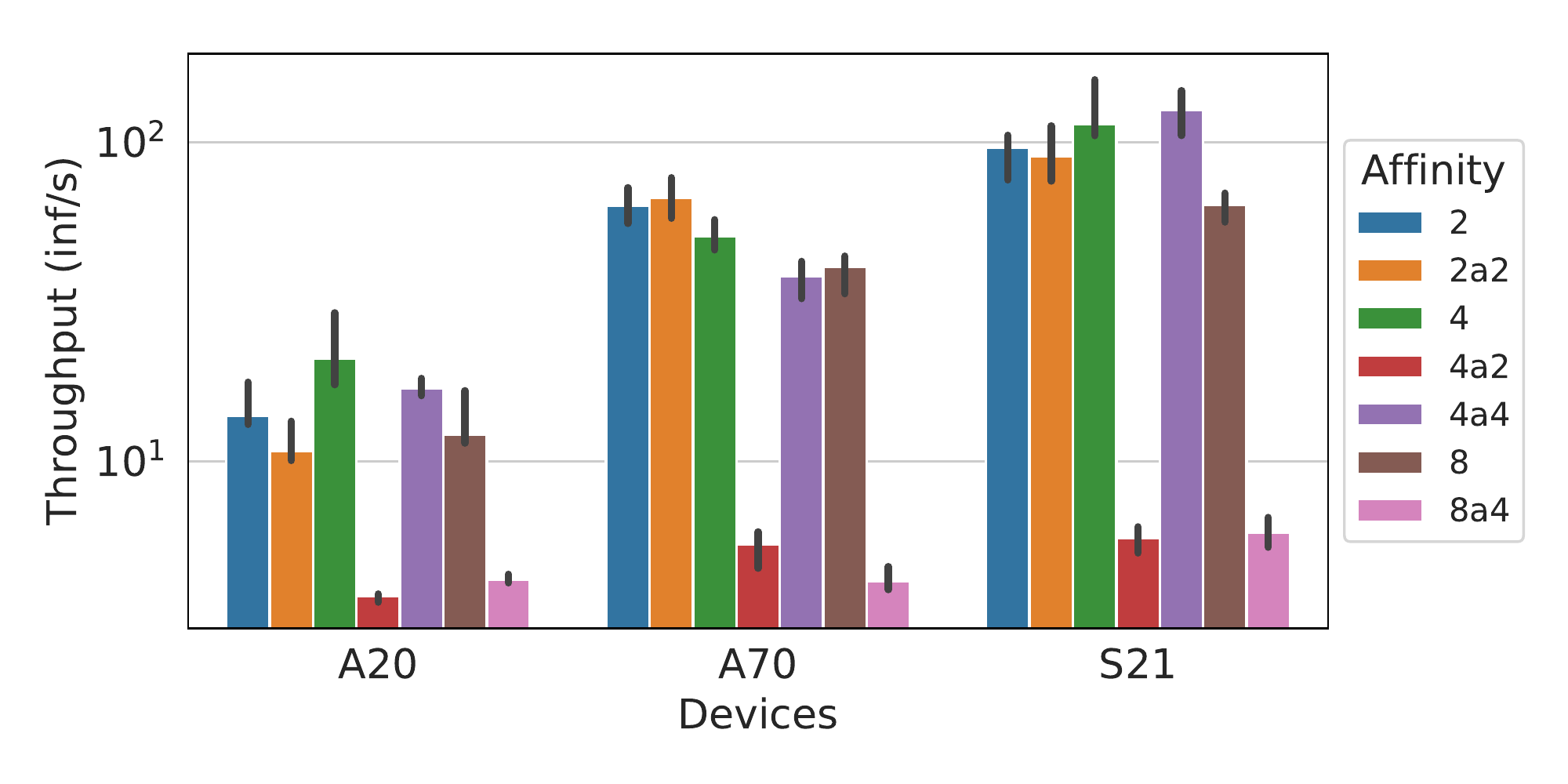}
    \vspace{-0.6cm}
    \caption{\text{TFLite}'s model throughput for different devices and compute targets.}
    \vspace{-0.5cm}
    \label{fig:threads}
\end{figure}

\noindent
\textbf{Quantisation:}
\label{sec:quantisation}
Finally, quantisation constitutes a prominent method for minimizing the computational and memory demands of DNNs by means of reducing their representation precision~\cite{qcnns2016cvpr,int_only2018cvpr}.
To study its adoption, we analysed the layer types and their weight and input bitwidth representations.
We report that 10.3\% of the models make use of the \texttt{dequantize} layer, which indicates the deployment of lower-precision models as a way to perform model compression. 
Furthermore, by examining each model's weights, we found that 20.27\% of the models use \texttt{int8} for the weight tensors whereas 10.31\% of the models work with \texttt{int8} activations.

Recent hardware advances have led to NPUs that support \textit{multiple arithmetic precisions}~\cite{snpe,arm_ethos_npu,huawei_npu2019hotchips}. Such examples are the Hexagon 698 processor on Qualcomm Snapdragon 865 (SDM865)~\cite{snpe} and the Arm Ethos processor~\cite{arm_ethos_npu}, which support 16-bit for activations and 8-bit for weights (A16W8).
These schemes enable a better compromise between faster low-precision compute and having enough representational power to achieve good accuracy.
In spite of the new opportunities of these hardware architectures, not only do existing deployment methodologies fail to exploit them but we also found no evidence of their adoption. We revisit the issue of quantisation with hardware-specific optimisations in Sec.~\ref{sec:hw-specific-optimisations}, where we use the Google's \texttt{NNAPI} and Qualcomm's \texttt{SNPE} to target specific processors in the SoC.

\noindent
\textbf{Observations:}
\emph{While the research community has developed numerous ways to optimise DNNs for mobile execution,
out-of-the-box support for such optimisations in modern frameworks' can be primitive and might not translate to run time gains at the expense of accuracy. Furthermore, most optimisations typically require model re-training and access to large-scale datasets.  As such, we find that such optimisations are not widely adopted by the mobile AI developers. Quantisation, which can also be used to target different SoC accelerators, is the most widely-used optimisation. However, more advanced hybrid quantisation schemes remain unsupported.}

\subsection{System-level optimisations}
\label{sec:system-level-optimisations}
Upon deploying a model, developers have different setup choices that can affect the model's performance.
In this section, we discuss the impact of different tuneable model and system parameters on model performance.

\noindent
\textbf{Impact of batch size.}
One common way of increasing a model's throughput is batching input samples together. By taking advantage of SIMD instructions of SoCs and accelerators, this technique increases the DNNs throughput by producing multiple inference results in one forward pass.

In Fig.~\ref{fig:batch}, we show the batch throughput across devices when processing $2, 5, 10,$ and $25$ samples at a time with 4 threads. We only consider \texttt{TFlite} models that successfully ran all batch sizes across all devices (149 in total). %
As expected, we see that the throughput increases as the batch size does. 
In fact, throughput scales almost linearly, which indicates that no bottleneck is hit up to that point.
Moving the comparison across devices, we see that S21 offers significantly faster inference, with throughput being $2.14\times$ and $5.42\times$ higher compared to A70 and A20 respectively on the highest batch size. This result goes in line with our conclusions from Sec.~\ref{sec:on-device-lat}. We anticipate that when scaling to higher batch sizes, devices with lower core count and memory will hit memory bandwidth bottlenecks or out of memory errors, but we defer this for future work.

\begin{figure}[t]
    \centering
    \includegraphics[width=\columnwidth]{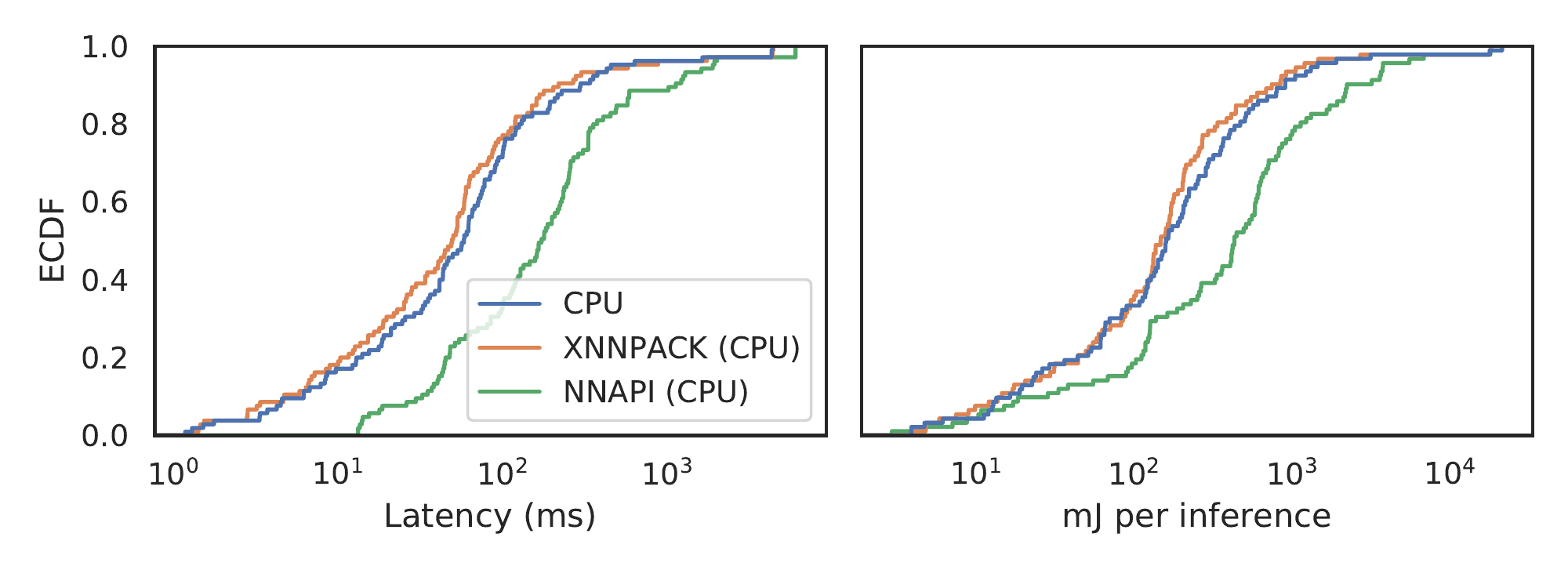}
    \vspace{-0.65cm}
    \caption{ECDF of TFLite models latency and energy per CPU runtime.}
    \vspace{-0.4cm}
    \label{fig:nnapi-xnnpack-ecdf}
\end{figure}

\noindent
\textbf{Impact of thread count.}
Another tuneable parameter during mobile execution is the number of threads allocated for execution on CPU. By default, all cores of the device can be \cready{simultaneously} used during execution (ARM DynamIQ). However, in Heterogeneous Multi-core Processors (HMP) there usually exist multiple islands of cores, offering different dynamics and computational power.
In Fig.~\ref{fig:threads} we show how the models' throughput varies when executed with different thread counts (2,4,8) and affinities (2,4). For the latter, we use process pinning to select which cores to target \cready{from} the heterogeneous core sets. 
We observe that the optimal thread count can vary across devices, with A20, A70 and S21 performing better with 4, 2 and 4 threads, respectively. 
We also see that the 8-threaded performance drops significantly across devices, indicating bottlenecked execution.

Digging deeper into thread performance, we further plot four additional setups where we set the CPU affinity to run over a varying number of the largest cores. For example, \textit{4a2} means 4 threads with affinity 2, which means 4 threads will run over the top 2 cores of the mobile's SoC.
As expected, we observe that any setup that sets the number of threads higher than the CPU affinity cores (\textit{4a2} and \textit{8a4}) results in significant performance degradation. This happens to due to time-sharing, having the other thread pinned on the same core waiting.
Nonetheless we also witness some less expected findings, such as the fact that setting the affinity to the same number of top cores does not yield any significant gain, irrespective of our initial hypothesis that it would reduce process migration between cores. In fact, \textit{4a4} performs worse to 4 \cready{threads for A70} and similar is the case for \textit{2a2} and 2 \cready{threads} for A20.

Predicting the optimal number of threads for mobile inference can be challenging as mobile devices have different CPU architectures with varying core frequencies as well as DVFS-enabled schedulers implementing energy-preserving policies \cite{kim2017enhancing}. Moreover, most mobile devices, nowadays, incorporate HMP SoCs (i.e. ARM big.LITTLE, DynamIQ) with varying number of cores per island (e.g. Q888 has $1\times$X1, $3\times$A78, $4\times$A55 ARM Cortex cores, whereas Q675 has $2\times$A76 and $2\times$A55 cores). Therefore, scheduling across core islands can bring sub-optimal results to DNN execution. However, when selecting the optimal thread count and affinity for each device, we see up to $2\times$ throughput gains overall.
This suggests that tuning scheduling and thread count of DNN execution on heterogeneous devices and processors can yield significant improvements.

\noindent
\textbf{Observations:}
\textit{Results from model-level optimisation indicate that there are alternative parameters for boosting inference throughput, but they should be tweaked in tandem with system-level factors, including the SoC topology and memory hierarchy to make efficient use of the underlying hardware.}

\begin{figure}[t]
    \centering
    \includegraphics[width=\columnwidth]{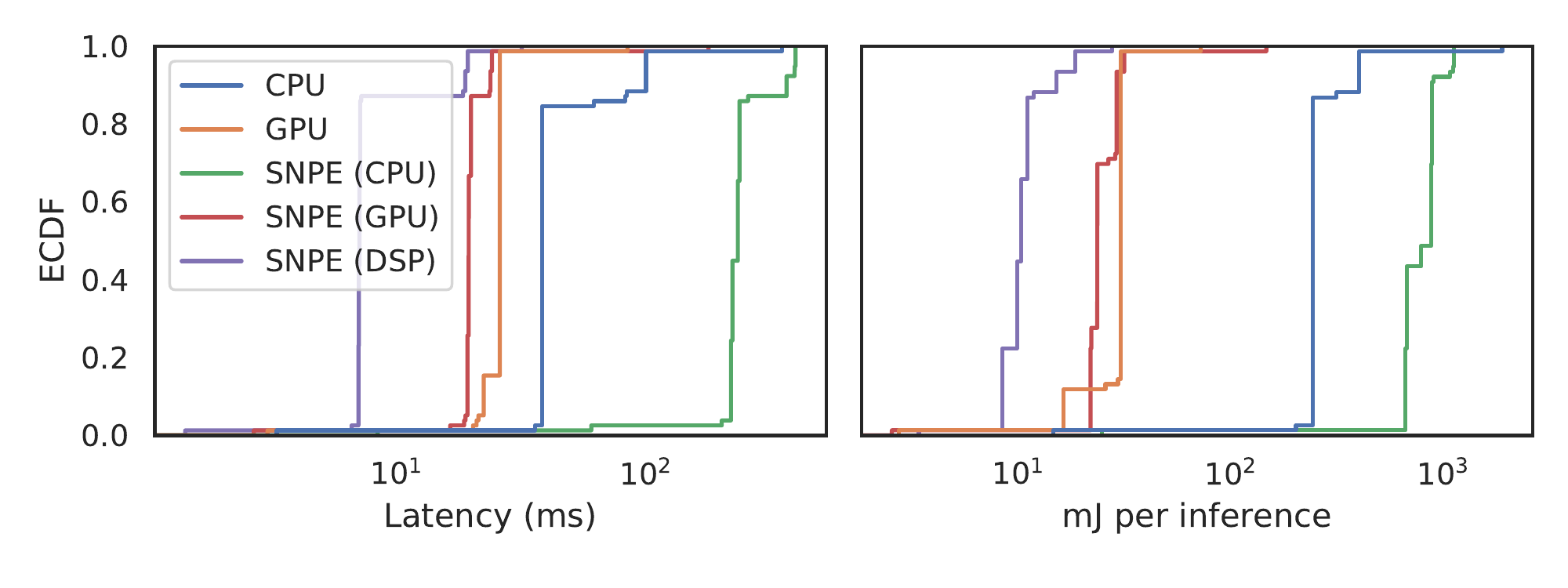}
    \vspace{-0.65cm}
    \caption{ECDF of TFLite and caffe models latency and energy per hardware target with SNPE.}
    \vspace{-0.4cm}
    \label{fig:snpe-ecdf}
\end{figure}

\subsection{Target generality vs. hardware-specific optimisations}
\label{sec:hw-specific-optimisations}

In the previous section, we have visited certain setup ``hyperparameters'', namely \textit{batch size} and \textit{process affinity} that depending on the use-case can enhance inference performance. In this section, we investigate framework-specific optimisations that can enhance performance, either by means of optimised operator kernel implementations or by moving computation to a different device altogether, i.e. targeting the GPU/NPU/DSP of the SoC. 
To this direction we run experiments measuring performance and energy of framework-specific optimisations on \texttt{TFLite} and \texttt{caffe} models across three alternative backends, namely \texttt{NNAPI}, \texttt{XNNPACK} and \texttt{SNPE}, on the Q845 board. We divert the reader to the Appendix for more information on these frameworks.

\noindent
\textbf{Traces of hardware-specific acceleration.} In our latest snapshot, we found some traces of hardware-specific acceleration. Specifically, we have found $71 (23.8\%)$ apps are using \texttt{NNAPI}, a single application using \texttt{XNNPACK} and three using \texttt{SNPE}. It is interesting to note that in the last case these models get blindly distributed to all devices, irrespective of having a Qualcomm-based SoC or not. In fact, they deploy both a \texttt{TFLite} and \texttt{dlc} variants of the same model.
Overall, we see that many app models are missing on the efficiency promises of targeting specialized hardware or using target-optimized kernel operations.

\noindent
\textbf{Optimisation opportunities.}
As a way to measure the potential benefit of using each of the aforementioned framework optimisations on different processing elements, we run two experiments, one on \texttt{TFLite} models for \texttt{NNAPI}  and \texttt{XNNPACK} (Fig.~\ref{fig:nnapi-xnnpack-ecdf}) and another for \texttt{TFLite} and \texttt{caffe} models for \texttt{SNPE} (Fig.~\ref{fig:snpe-ecdf}). In each case, we compare the performance of framework-specific optimisations to the baseline CPU and GPU runs. The reason we do not compare across them is because the number of models commonly compatible is low. This highlights one succinct characteristic of such optimisations, the rudimentary support for operators across heterogeneous targets which in turn can hinder their widespread adoption.

Results from our evaluation indicate that for CPU execution (Fig.~\ref{fig:nnapi-xnnpack-ecdf}), one is better off using the XNNPACK delegate for executing DNN inference $1.03\times$ faster and $1.13\times$ more efficiently on average. \texttt{NNAPI} did not prove its potential in our experiments, with its performance lagging behind the default CPU execution ($0.49\times$ slower and $1.66\times$ less efficient on average). This could be potentially attributed to unoptimised NN drivers from the vendor.
On the other hand, when one is deploying with a vendor-specific platform, \texttt{SNPE} in our case, performance is better for DSP and GPU (Fig.~\ref{fig:snpe-ecdf}), compared to vanilla CPU and GPU runs. Specifically, these are $5.72\times$ and $2.28\times$ faster and $20.3\times$ and $8.39\times$ more efficient on average, compared to CPU runs. In comparison to GPU runs, these are $2.97\times$ and $1.19\times$ faster and $2.69\times$ and $1.11\times$ more efficient on average. In the case of CPU, however, the story is similar with our last experiment, further corroborating the story for non-optimised CPU drivers from the vendor.

Note that CPU and GPU runs are executed at full-precision (\texttt{float32}), while the DSP runs in \texttt{int8}. Depending on the task this can result in accuracy variations, but we do not have access to model-specific data and labels to assess that.

\noindent
\textbf{Observations.}
\textit{Results from our experiments say a mixed story about hardware and frameworks specific optimisations. While it can yield noticeably better performance across models, this is not always the case due to driver implementation or other low-level confounding factors. The dilemma of target generality vs hardware-specific optimisations ultimately lies in the hand of the developer and the resources they have at their disposal to extract every bit of performance in hardware.}

\vspace{-0.2cm}
\subsection{Cloud-based DNN models}

Another approach to accelerate inference and bring intelligence to mobile apps, without having the need to specialise per target device is by offloading to the cloud.
We can envision this approach being popular amongst developers who do not implement or train their own models or for models that are too computationally intensive to run locally on a mobile device or too expensive to optimise for each available target to offer a similar QoE.

\begin{figure}[t]
    \centering
    \includegraphics[width=0.75\columnwidth]{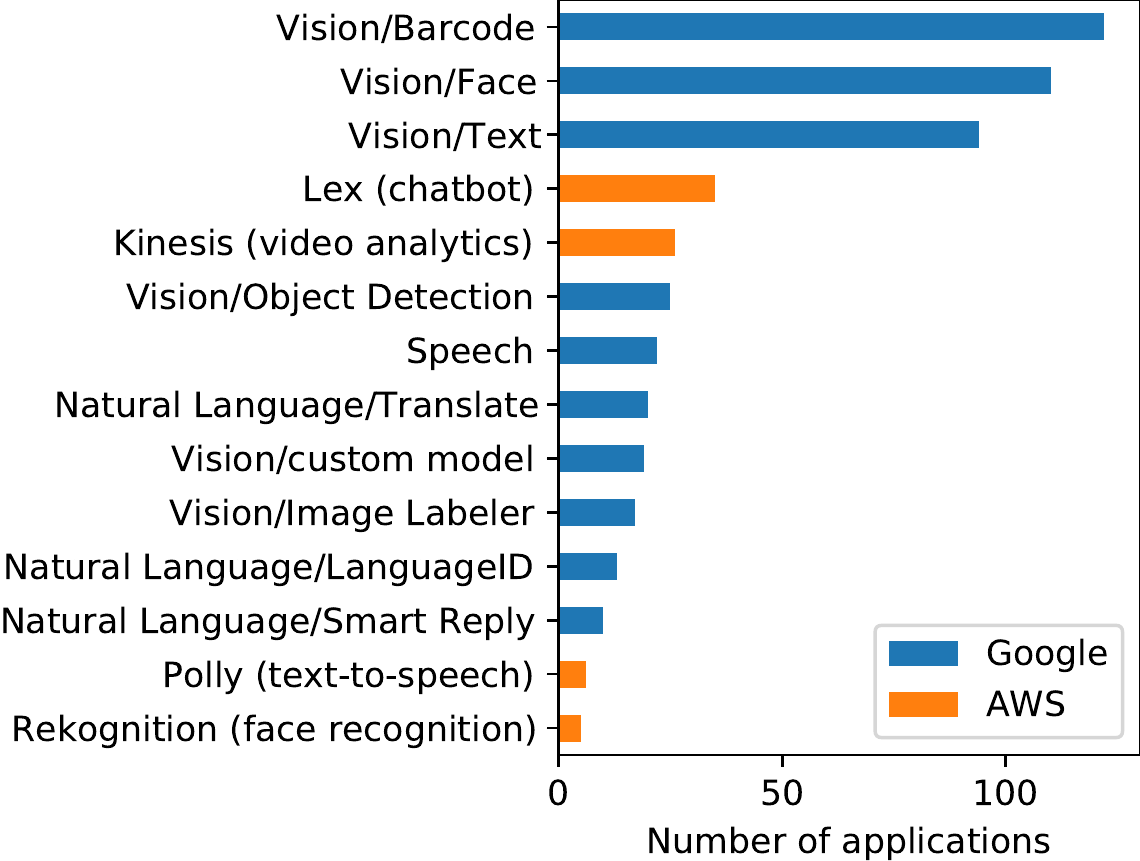}
    \vspace{-0.2cm}
    \caption{\cready{Number of apps that invoke cloud-based ML APIs. Categories with less than 10 apps are excluded.}}
    \vspace{-0.4cm}
    \label{fig:cloud}
\end{figure}

As mentioned in Sec.~\ref{sec:methodology_offline}, \tool{} tracks app invocations of known cloud-based machine learning APIs in their code. This includes calls to Google (Google Cloud and Firebase ML) and Amazon services. 
\cready{Fig.}~\ref{fig:cloud} \cready{shows the number of applications invoking each of the cloud-based ML APIs across our dataset.}
Overall, we find 524 distinct applications that use cloud AI APIs, a considerable increase of $2.33\times$ from our 2020 dataset.
More specifically, 452 and 72 apps using Google AI services and Amazon respectively.
\cready{This increase is inline with the increase in models deployed within the apps }(Sec.~\ref{sec:temporal}).
\cready{Furthermore,} we observe that developers primarily use cloud-based image and video analytics to perform face identification, bar/QR code recognition, video analytics and chatbots.

\noindent
\textbf{Observations:}
\emph{Our results indicate that cloud APIs from Google and Amazon are gaining in popularity as they allow developers to quickly deploy AI capabilities without the need for specialised ML expertise and costly infrastructure for training. Moreover, developers do not need to maintain training data on-premise and the resulting apps can be supported by heterogeneous devices with similar QoE.}

\section{Related Work}
\label{sec:related_work}

In the past, there have been numerous studies that performed large scale analysis of the Google Play Store but with different aims, such as characterising mobile apps~\cite{viennot2014measurement} and their API usage~\cite{onwuzurike2018family,alarms}.
Closer to the ML community, there has been an increasing effort to benchmark state-of-the-art models across different devices and frameworks~\cite{mmsys2018,ase2019,fb_edge2019hpca,characterizing2019iiswc, almeida2019embench,ai_benchmark_2019}. 
Although these studies have done a great job at extensively benchmarking state-of-the-art models, we still lack the knowledge as to whether these models are representative of the ones deployed today in mobile apps. 
Moreover, there is a lack of understanding on how the latest trends on DNN optimisation affect the latest DNN-based mobile apps. 

To the best of our knowledge, there are largely two works that have investigated DNN usage in the wild. One is from  Xu et al.~\cite{xu2019first} and focuses on investigating who the early adopters of DNNs are and what are the use-cases for Deep Learning in mobile apps. While they do conduct a lightweight analysis of DNN operations, they have only measured model footprint and performance in an offline and device-agnostic manner, by means of measuring the FLOPs of DNN layers. However, it has been shown that FLOPs is not a good proxy of a model's run time ~\cite{almeida2019embench,ai_benchmark_2019}, especially across different hardware configurations. Therefore, there is still limited understanding about the actual performance of DNN models in the wild, across a heterogeneous ecosystem of more and less capable devices.
A more privacy-centric work has been presented in \cite{sec_dnns_apps}, which investigates DNN model protection on mobile devices and illustrates succinctly that many Android apps do not protect their DNN models, which means these can be easily leaked, or extracted for analysis. Nevertheless, it does not perform any performance analysis.

These two works serve as a starting point for our study, which aims to answer the question of how widely deployed DNNs found in the most popular Android apps actually perform on widely deployed devices, essentially correlating the state of Deep Learning mobile deployment in the wild. To this end, we conduct an in-depth benchmarking of models used in the latest most trending mobile apps. This includes analyses of latency, energy, system and model-level parameters and optimisations, providing a better comprehension of the current limitations when deploying DNNs on mobile phones of different tiers and generations.

\section{Discussion \& Future work}
\label{sec:discussion}
\subsection{Implications \& Trends}

\noindent
\textbf{Proliferation of mobile AI.} Our results indicate that both on-device and cloud-supported DNN applications are increasing rapidly (doubled within a year). This is mostly driven by the availability of pre-trained models and easy-to-use cloud-based APIs, focusing mostly on vision tasks such as image detection and recognition. 

\noindent
\textbf{Model reuse.} While there is \cready{much} research on bespoke model architectures, customisation and fine-tuning \cite{pan2009survey,emdl_ee_survey}, we observe that most developers use off-the-shelf DNN architectures.
In fact, close to 80.9\% of the models are shared across two or more applications and a further 9.02\% of the remaining models share some layers (i.e., derived from a common model after fine-tuning). 
Simultaneously, there is a parallel trend of resorting to  cloud-powered inference, further demonstrating  a preference of developers towards turnkey solutions, instead of bespoke customised %
\cready{ones}. With the current trajectory of AI, we expect more developers specialising on ML-based app development at least until the middleware (e.g. \texttt{NNAPI}) \cready{which} abstracts away ML-specific parameters becomes more prevalent.

\noindent
\textbf{DNNs and mobile hardware resources.}
We witness that most applications do not take advantage of SoC-specific accelerators to accelerate their inference runtime, but rather target generality of their solutions, either by shipping vanilla CPU-only execution or by integrating framework-specific middleware options (e.g. \texttt{NNAPI}). Last, offloading inference to the cloud offers a consistent QoE, which is not dependent on the target device, at the expense of privacy \cite{spinn2020mobicom, almeida2021dyno} and monetary cost.
This behaviour comes as a consequence of the fragmentation in the Android ecosystem in terms of hardware capabilities and software support (e.g. vendor-specific \texttt{NNAPI} drivers).
Consequently, we anticipate the need of automated solutions for optimised development and deployment of ML solutions in mobile apps, which abstract away the complexity of efficiency and heterogeneity of the ecosystem.

\noindent
\textbf{Energy as a bottleneck.}
While Deep Learning adoption is undisputed, with accelerating trajectory in the future, manufacturers turn to specialised hardware for faster and more efficient ML (e.g. NPUs). However, the same cannot be stated for battery technology and capacity, which remain relatively stale. 
Given what we observed for the segmentation scenario in Sec.~\ref{sec:energy-scenarios}, we anticipate energy sooner or later becoming a bottleneck in DNN deployment, requiring novel solutions to support mobile intelligence on the go. 

\noindent
\textbf{DNN co-habitation.}
With more and more applications shipping DNN-powered solutions, we also anticipate the co-existence and parallel runtime of more than one DNN in the future. Thus, researchers will need to tackle this emerging problem to efficiently support such runtimes, by means of OS or \mbox{hardware-level solutions.}

\noindent
\textbf{On-device learning and personalisation.}
Last, so far in the paper we have only visited the task of mobile inference.
In this setup, the weights of the model come pretrained on some centralised dataset and the device only performs forward propagation. However, with users becoming more and more privacy aware and with legislation discouraging the storage of user data without legitimate interest, on-device training and federated learning \cite{mcmahan2017communication,horvath2021fjord} become more and more prevalent \cite{paulik2021federated,MLSYS2019_bd686fd6}.
Moreover, with the proliferation of on-device data, on-device personalisation \cite{10.1145/3446382.3448359} is also gaining traction.
These tasks will create a different workload to be optimised for on-device runtime, for which current or future tools will need to provide support.

\vspace{-0.2cm}
\subsection{Limitations}
In this work we have shed light to the use and performance of DNNs in real-world applications. However, we only focused on the Android smartphone landscape due to its larger market share and wide device fragmentation. These finding might only partially hold for other mobile ecosystems. %

Furthermore, we have analysed the models that could be identified as DNN models. Obfuscated and encrypted models, or models that are downloaded outside of Google Play store were not benchmarked, despite us tracking the respective application as ML-powered. While there might be a different distribution of obfuscated models in the wild, the results from \cite{sec_dnns_apps} indicate otherwise.

Our analysis included both offline introspection and dynamic benchmarking of the models. 
However, we did not investigate particular invocation paths and frequency of inference per app. We expect that some of these models are rarely used (e.g. credit card scanning) while others are utilised more frequently (e.g. activity detection). 
However, the real-world usage of these models requires device instrumentation and collecting telemetry data over a large user-base. %
\cready{While previous works~\cite{almeida2018chimp,onwuzurike2018family} have proposed large-scale crowd-testing of virtualised mobile apps with real user interaction, these generally preclude testing sensor input-dependent functionality, on which DNNs depend.}
We leave this as future work.

Last, while we characterise DNN cloud offloading, we acknowledge that we miss any developers who use their own custom (e.g., REST-based) APIs to access remote execution.

\section{conclusion}

In this work, we have carried out a comprehensive empirical study of the most popular DNN-powered mobile apps. 
Using \tool{}, we analyse thousands of mobile apps in the wild and identify a significant chasm between the deployed models and the state-of-the-art architectures and optimisation techniques. 
This is the first work to dig deeper into these aspects so as to provide guidelines for both the mobile application and the DNN-framework developer communities.

\bibliographystyle{ACM-Reference-Format}
\bibliography{acmart}

\newpage
\clearpage

\appendix

\section{Additional platform information}

\subsection*{DNN Model extraction}
\label{sec:A-formats}

In Sec.~\ref{sec:crawling} of the paper, we stated that \tool{} supports file extraction from i)~the base \texttt{apk}, ii)~expansion files (\texttt{OBB}s) and iii)~Android App Bundles.
The extracted files are matched against a compiled list of known DNN framework formats and validation rules to identify potential DNN models. The complete list of formats is shown in Table~\ref{tab:support}.

\begin{table}[h!]
\resizebox{0.99\linewidth}{!}{%
\begin{tabular}{ll}
\hline
Framework & Extensions \\ \hline
ONNX & .onnx, .pb, .pbtxt,  .prototxt \\ \hline
MXNet & .mar, .model, .json,  .params \\ \hline
Keras & .h5, .hd5, .hdf5,  .keras, .json,  .model, .pb, .pth \\ \hline
Caffe & .caffemodel, .pbtxt,  .prototxt, .pt \\ \hline
Caffe2 & .pb, .pbtxt, .prototxt \\ \hline
PyTorch & .pt, .pth, .pt1, .pkl, .h5, .t7, .model, .dms, .pth.tar, .ckpt,  .bin, .pb, .tar \\ \hline
Torch & .t7, .dat \\ \hline
SNPE & .dlc \\ \hline
FeatherCNN & .feathermodel \\ \hline
TFLite & .tflite, .lite, .tfl, .bin, .pb \\ \hline
TF & .pb, .meta, .pbtxt, .prototxt, .json, .index, .ckpt \\ \hline
Sklearn & .pkl, .joblib, .model \\ \hline
armNN & .armnn \\ \hline
Mnn & .mnn \\ \hline
Ncnn & .param, .bin, .cfg.ncnn, .weights.ncnn, .ncnn \\ \hline
Tengine & .tmfile \\ \hline
Flux & .bson \\ \hline
Chainer & .npz, .h5, .hd5, .hdf5, .chainermodel \\ \hline
\end{tabular}}
\caption{Frameworks and formats validated by \tool{}}
\label{tab:support}
\end{table}

\section{Additional experiment information}

\subsection*{Hardware-specific acceleration frameworks}

As per Sec.~\ref{sec:hw-specific-optimisations}, we run our \texttt{TFLite} models against alterative backends, namely \texttt{NNAPI}, \texttt{XNNPACK} and \texttt{SNPE}. Below we provide additional information for each one:

\begin{itemize}[leftmargin=0pt,topsep=0pt,label={}]
    \item \textbf{NNAPI}\footnote{\url{https://developer.android.com/ndk/guides/neuralnetworks}}\textbf{.} Neural Networks API (\texttt{NNAPI}) is a middleware-level library in Android that sits between the machine learning framework library used by an application (e.g. \texttt{TFLite}) and the the Android Hardware Acceleration Layer (HAL). It essentially provides an abstraction layer, handling hardware acceleration through vendor and hardware specific NN drivers, which provide efficient operator implementations for CPU, GPU, DSP, NPUs or other kinds of specialised hardware. Execution falls back to CPU in the absence of such drivers or unsupported operators. \texttt{TFLite} is at the foreforent of NNAPI delegation and PyTorch Mobile has announced support for it. Nonetheless, \texttt{NNAPI} being in its infancy comes with some shortcomings, mainly in the realm of OS version support (Android P and above), NN drivers availability and heterogeneity in performance gains.
    \item \textbf{XNNPACK}\footnote{\url{https://github.com/google/XNNPACK}}\textbf{.} \texttt{XNNPack} provides a low-level, highly optimised library for NN inference operators across platforms. Specifically for ARM, it supports efficient implementation of operators through Neon instructions, as well as inference on sparse networks, which offers a practical solution to the problem described in Sec.~\ref{sec:pruning}. Despite the claimed performance benefits, operator support is limited and if not careful can lead to performance penalties instead of gains when compared to the baseline CPU delegates.
    \item \textbf{SNPE}\footnote{\url{https://developer.qualcomm.com/docs/snpe/overview.html}}\textbf{.} The Snapdragon Neural Processing Engine (\texttt{SNPE}) constitutes a vendor-specific runtime for execution of DNNs on Qualcomm SoCs, targeting the CPU, Adreno GPU or Hexagon DSP of the SoC, handling quantisation in the proper precision internally. It uses its own representation for NNs (\texttt{.dlc} format) supports conversion from different frameworks, including \texttt{caffe} and \texttt{TFLite}. However, while SNPE can potentially take advantage of hardware-specific optimisations, it can only target Qualcomm SoCs, trading off generality for performance. Operator support can also be of issue in \texttt{SNPE}, supporting CPU fallback in case of hardware-specific unsupported operations.
\end{itemize}

\end{document}